\documentclass[pdflatex, default, iicol]{sn-jnl}

\jyear{2021}%

\usepackage[english]{babel}
\usepackage{times} 
\usepackage[utf8]{inputenc}
\usepackage{hyperref}
\usepackage[nolist]{acronym}
\usepackage{balance} 

\DeclareMathAlphabet{\pazocal}{OMS}{zplm}{m}{n}

\makeatletter
\providecommand{\bigsqcap}{%
	\mathop{%
		\mathpalette\@updown\bigsqcup
	}%
}
\newcommand*{\@updown}[2]{%
	\rotatebox[origin=c]{180}{$\m@th#1#2$}%
}
\makeatother

\usepackage{siunitx}
\usepackage{graphicx} 
\usepackage{epsfig} 
\usepackage{caption}
\usepackage{subcaption}

\usepackage[export]{adjustbox}

\makeatletter
\let\MYcaption\@makecaption
\makeatother

\makeatletter
\let\@makecaption\MYcaption
\makeatother

\usepackage{mathtools}
\usepackage{nicefrac}
\usepackage{amsmath}
\usepackage{amssymb}
\usepackage{bm} 

\def\x#1{\texttt{\expandafter\string\csname#1\endcsname}&\expandafter$\csname#1\endcsname$}

\usepackage{etoolbox}
\makeatletter%
\AfterPreamble{%
	\usepackage{hyperref}%
	\let\oldhypertarget\hypertarget%
	\renewcommand{\hypertarget}[2]{%
		\oldhypertarget{#1}{#2}%
		\protected@write\@mainaux{}{%
			\string\expandafter\string\gdef%
			\string\csname\string\detokenize{#1}\string\endcsname{#2}%
		}%
	}%
	\newcommand{\myhyperlink}[1]{%
		\hyperlink{#1}{\csname #1\endcsname}%
	}%
}
\makeatother%

\newcounter{Definition}
\newcommand{\displayDefinitions}[2][]{%
	\stepcounter{Definition}%
	\textit{Definition~}\hypertarget{#1}{\theDefinition}\textit{~(#2)}%
}

\newcommand{\refDefinitions}[1][]{%
	Def.~\myhyperlink{#1}%
}

\newcounter{Problem}
\newcounter{Theorem}
\newcounter{Proposition}
\newcounter{Remark}
\DeclareMathOperator*{\minimize}{minimize}
\DeclareMathOperator*{\maximize}{maximize}

\usepackage{xcolor}
\definecolor{myYellow}{rgb}{0.93,0.69,0.13}
\definecolor{myPurple}{rgb}{0.49,0.18,0.56}

\begin{document}

\title[A STL Motion Planner for Bird Diverter]{A Signal Temporal Logic Motion Planner for Bird 
Diverter Installation Tasks with Multi-Robot Aerial Systems}


\author[1]{\fnm{Alvaro} \sur{Caballero}}\email{alvarocaballero@us.es}
\equalcont{These authors contributed equally to this work.}

\author*[2]{\fnm{Giuseppe} \sur{Silano}}\email{giuseppe.silano@fel.cvut.cz}
\equalcont{These authors contributed equally to this work.}



\affil[1]{\orgdiv{Department of System Engineering and Automation}, \orgname{University of Seville}, \orgaddress{\street{Camino de los Descubrimientos s/n}, \city{Seville}, \postcode{41092},  \country{Spain}}}

\affil*[2]{\orgdiv{Department of Cybernetics}, \orgname{Czech Technical University in Prague}, \orgaddress{\street{Karlovo Namesti 13}, \city{Prague}, \postcode{12135}, \country{Czech Republic}}}



\begin{acronym}
    \acro{CBF}[CBF]{Control Barrier Function}
    \acro{CVRP}[CVRP]{Capacitated Vehicle Routing Problem}
    \acro{CNN}[CNN]{Convolutional Neural Network}
    \acro{FOV}[FoV]{Field of View}
    \acro{ICP}[ICP]{Iterative Closest Point}
    \acro{ILP}[ILP]{Integer Linear Programming}
    \acro{IP}[IP]{Initial Position}
    \acro{GNSS}[GNSS]{Global Navigation Satellite System}
    \acro{GPS}[GPS]{Global Positioning System}
    \acro{LSE}[LSE]{Log-Sum-Exponential}
    \acro{MILP}[MILP]{Mixed-Integer Linear Programming}
    \acro{MPC}[MPC]{Model Predictive Control}
    \acro{ML}[ML]{Machine Learning}
    \acro{NLP}[NLP]{Nonlinear Programming}
    \acro{NN}[NN]{Neural Network}
    \acro{PRM}[PRM]{Probabilistic Roadmap}
    \acro{RRT}[RRT]{Rapidly-exploring Random Tree}
    \acro{ROS}[ROS]{Robot Operating System}
    \acro{ROW}[ROW]{Rolling On Wire}
    \acro{RS}[RS]{Refilling Station}
    \acro{SIL}[SIL]{Software-in-the-loop}
    \acro{SST}[SST]{Stable Sparse RRT}
    \acro{STL}[STL]{Signal Temporal Logic}
    \acro{TL}[TL]{Temporal Logic}
    \acro{TR}[TR]{Target Region}
    \acro{TSP}[TSP]{Traveling Salesman Problem}
    \acro{UAV}[UAV]{Unmanned Aerial Vehicle}
    \acro{UGV}[UGV]{Unmanned Ground Vehicle}
    \acro{VRP}[VRP]{Vehicle Routing Problem}
    \acro{wrt}[w.r.t.]{with respect to}
\end{acronym}



\abstract{%
This paper addresses the problem of task assignment and trajectory generation for installing bird diverters using a fleet of multi-rotors. The proposed solution extend our previous motion planner to compute feasible and constrained trajectories, considering payload capacity limitations and recharging constraints.~\ac{STL} specifications are employed to encode the mission objectives and temporal requirements. Additionally, an event-based replanning strategy is introduced to handle unforeseen failures. An energy minimization term is also employed to implicitly save multi-rotor flight time during installation operations. The effectiveness and validity of the approach are demonstrated through simulations in MATLAB and Gazebo, as well as field experiments carried out in a mock-up scenario.
}



\keywords{%
Aerial Systems: Applications, Formal Methods in Robotics and Automation, Multi-Robot Systems, Task and Motion Planning. 
}


\maketitle


\section{Introduction}
\label{sec:introduction}

\begin{figure}[tb]
    \centering
    \includegraphics[width=\columnwidth]{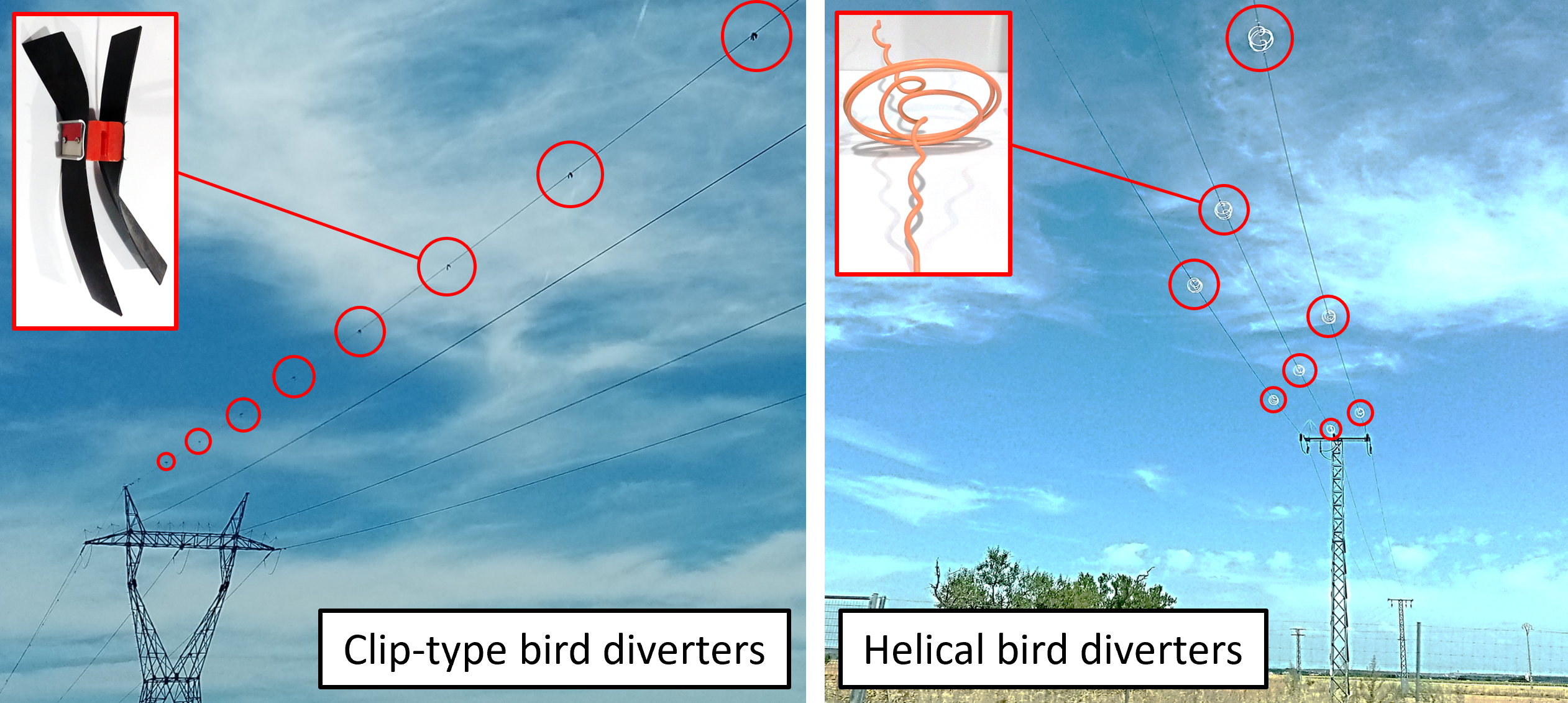}
    \caption{Various types of bird diverters installed on a medium voltage power line infrastructure~\cite{Suarez2021Access}}
    \label{fig:typeDiverters}
\end{figure}

Power lines play a crucial role in providing energy to millions of people and are considered vital civil infrastructure in any country. To enhance network reliability and minimize power outages, electricity supply companies invest significant resources in inspection and maintenance operations~\cite{EPRI2008TechnicalReport}. Among these activities, the installation of bird diverters on power lines (see Figure~\ref{fig:typeDiverters}) is essential to mitigate the risk of bird collisions~\cite{Hunting2002CECOM} and improve their visibility~\cite{Ferrer2020GEC}. Bird mortality caused by power line collisions is a significant concern, particularly in areas with diverse bird populations or during migratory seasons. It has been estimated that power line collisions affect approximately 350 bird species~\cite{Manville1999}, resulting in millions of bird deaths annually worldwide~\cite{Manville2008, Jenkins2010BCI}.

To address this issue, various types of bird diverters have been developed, including active and passive designs. \textit{Active bird diverters} utilize wind-driven components, while \textit{passive diverters}, such as helical objects made of plastic or aluminum, are attached to power cables to serve as visual markers (see Figure~\ref{fig:typeDiverters}). Additionally, alternative techniques, such as visual and auditory deterrents, have been developed to mitigate bird collisions. \textit{Visual deterrents} employ markers or reflective materials to enhance visibility and assist birds in avoiding power lines, while \textit{auditory deterrents} emit sounds that discourage birds from approaching. However, the effectiveness of these techniques can vary depending on bird species and local conditions. Integrating these techniques with other mitigation strategies provides a multi-faceted approach to enhance bird safety and maintain a reliable power supply~\cite{Ferrer2020}. However, the current method of using manned helicopters and experienced crews for the installation of bird diverters has drawbacks. Firstly, it is time-consuming, as power lines are often located in difficult-to-access areas. Secondly, it poses safety risks due to installations being performed at heights on active lines. 

\acp{UAV} offer a promising solution to automate bird diverter installation~\cite{Suarez2021AIRPHARO, Castano2021AppSci, CacaceDrones2023}. They can operate continuously over long distances and be equipped with lightweight manipulation devices for autonomous installation operations~\cite{Suarez2021Access, Armengol2021AIRPHARO, Suarez2021AIRPHARO, Afifi2022ICRA, Afifi2023ICUAS}. However, the limited battery and payload capacity of individual~\acp{UAV} necessitate the use of multiple~\acp{UAV} to expedite the process, respond to unforeseen events and cover large-scale scenarios. Planning for a multi-\ac{UAV} team presents challenges such as scheduling battery recharging and diverter installation, ensuring collision-free trajectories, considering vehicle dynamics and energy consumption models, among other concerns.

Therefore, advanced task and motion planning techniques are required to enable bird diverter installation using multi-\ac{UAV} teams while meeting safety requirements and mission objectives.~\ac{TL} can serve this purpose by providing a mathematical framework that combines natural language commands with temporal and Boolean operators~\cite{donze2010ICFMATS, Pola2019ARC}. In particular,~\acf{STL}~\cite{maler2004FTMATFTS} is equipped with a metric called \textit{robustness}, which not only evaluates whether the system execution meets requirements but also quantifies the extent to which these requirements are fulfilled. This leads to an optimization problem that aims to maximize the robustness score, thereby providing the best feasible trajectory while satisfying the desired specifications.

This paper proposes a motion planner for multi-\ac{UAV} systems that leverages~\ac{STL} specifications for the installation of bird diverters on power lines. The mission requirements are encoded as an~\ac{STL} formula, and a nonlinear non-convex max-min optimization problem is formulated to maximize the robustness score. To handle the complexity of this nonlinear optimization, a hierarchical approach is adopted. First, a~\ac{MILP} problem is solved, and the resulting solution is then fed into the final~\ac{STL} optimizer.



\subsection{Related work}
\label{sec:relatedWork}

This paper addresses a complex multi-\ac{UAV} motion problem that requires considering vehicle dynamics, collision avoidance, limited mission time, and payload capacity. Sampling-based planners~\cite{karaman2011IJRR} like the well-known~\ac{RRT} and its variants can compute collision-free trajectories in high-dimensional configuration spaces within reasonable time. Extensions have been proposed to minimize energy consumption and handle vehicle dynamics constraints~\cite{hauser_tro16, Webb2013ICRA}, although their convergence is not proven. However,~\ac{RRT}-based methods do not inherently handle payload capacity restrictions and struggle with complex nonlinear problems or multi-robot settings. To achieve real-time performance, motion primitives have been explored for optimal trajectory planning of~\acp{UAV}~\cite{Ryll2019ICRA, chen2021RAL}. Additionally, the problem often transforms into a combinatorial optimization challenge, where an exhaustive search becomes computationally infeasible. In such cases, heuristic methods are commonly employed to simplify the complexity and identify the most viable solution~\cite{Tan2021IEEEAccess, Wang2023AI, Wang2023OceanEngineering, Gao2023AI}.

When the mission requires visiting multiple locations, optimization approaches based on the~\ac{VRP} formulation have been proposed~\cite{Nekovar2021RAL, Xiong2022CCDC}. Since~\ac{VRP} variants are NP-hard combinatorial problems, many works propose heuristic approaches sacrificing optimality~\cite{coutinho2018unmanned, FaiglRAL2019}. While these methods can handle~\ac{UAV} capacity and mission time constraints, they face challenges in obstacle avoidance and multi-vehicle dynamics~\cite{penicka2019RAL}. Hierarchical approaches utilizing~\acp{PRM} and kinodynamic~\ac{SST} methods have been investigated to generate collision-free and minimum-time flight trajectories~\cite{Penicka2022RAL-I, Penicka2022RAL-II}. However, the problems discussed do not consider~\ac{UAV} payload capacity or multi-\ac{UAV} settings.

Alternatively, with the advancement of computational resources onboard~\acp{UAV} and recent progress in efficient numerical methods, receding horizon approaches for optimal control have emerged as a viable solution for addressing multi-\ac{UAV} trajectory planning~\cite{nguyen2021ECC}. Centralized methods for non-convex optimization~\cite{robinson2018RAL} and distributed~\ac{MPC}-based algorithms~\cite{luis2020RAL, alcantara2021RAS} have been proposed. Others have employed knowledge transfer techniques between agents to achieve collision avoidance and safety~\cite{Zhu2019RAL}. While these methods efficiently compute safe and optimal trajectories for multiple~\acp{UAV}, they do not incorporate high-level mission specifications (e.g., time interval or ordering constraints) or task planning into the problem formulation.

Moreover, literature includes publications that enhance motion planning for multi-robot systems using formal specification languages~\cite{maler2004FTMATFTS, CohenSpringer2023}. The authors in~\cite{Chen2012TRO} propose extracting local specifications assigned to specific robots to address multi-robot trajectory planning. Motion capabilities of each robot are represented using a transition system, which may become computationally expensive as the number of agents and tasks increase. On the other hand,~\acp{CBF}~\cite{Lindemann2020TCNS} have shown promise in efficient motion planning, but they are limited to a fragment of the~\ac{STL} formula, restricting their application to simple scenarios~\cite{Gilpin2021LCSS}. In~\cite{Leahy2021TRO}, the authors present robust algorithms for multi-robot coordination that encode high-level temporal logic specifications into an~\ac{MILP} problem. However, their sequential multi-robot~\ac{RRT} algorithm leads to suboptimal trajectories. A reinforcement learning approach for multi-agent systems has been proposed in~\cite{Muniraj2018CDC}, focusing on finite abstractions and deterministic systems. Nevertheless, methodological advances are still needed as it poses computational challenges for complex~\ac{STL} formulae. In~\cite{RamanCDC2014}, the authors propose a mathematical programming-based method for~\ac{MPC} using~\ac{STL} specifications. Nonetheless, its scalability is limited due to exponential complexity in the number of binary variables. In~\cite{Kantaros2020IJRR}, a scalable algorithm for multi-agent optimal control with~\ac{TL} is presented, but it lacks the ability to handle concrete timing requirements and agent counting. 



\subsection{Contributions}
\label{sec:contributions}

This paper introduces an innovative approach to task and motion planning for the installation of bird diverters on power lines using a team of multi-rotor~\acp{UAV}. The proposed method leverages~\ac{STL} to generate optimal trajectories that are dynamically feasible, considering vehicle dynamics and velocity and acceleration limits. These trajectories fulfill various mission specifications, including collision avoidance between~\acp{UAV} and the environment, adherence to~\ac{UAV} payload capacity limits, and reaching all installation targets within the given mission time while ensuring sufficient time for diverter installation and refilling operations. The problem results in a complex nonlinear non-convex max-min optimization problem, which is addressed through a novel hierarchical approach. 

Building upon our previous research~\cite{Silano2021RAL}, the work is extended to specifically address the installation of bird diverters, incorporating payload capacity limitations and considering the presence of refilling stations for long-endurance operations. However, finding an optimal solution in a reasonable time frame can be challenging due to the solvers' tendency to get stuck in local optima based on the initial guess~\cite{Bertsekas2012Book, CaiMacroeconomicDynamics2017}. Therefore, the work~\cite{Silano2021RAL} is also extend by introducing a novel procedure to compute the initial guess solution for the optimization problem. Additionally, our approach incorporates an event-based replanning strategy to handle unforeseen failures. Whenever a~\ac{UAV} fails, a new plan is computed online using the motion planner for a backup drone. This ensures uninterrupted mission progress. Lastly, to implicitly save multi-rotor flight time during installation operations, the proposed method also integrates an energy minimization term. The core idea is to determine an optimal cruising speed for each~\ac{UAV}, minimizing energy consumption, and generating trajectories that closely follow these nominal speeds. The main contributions of this paper are summarized as follows:

\begin{itemize}

    \item The multi-\ac{UAV} planning problem for installing diverters on power lines is formulated in Section~\ref{sec:problemDescription}. We extract~\ac{STL} specifications (see Section~\ref{sec:preliminaries}) and employ them to establish an optimization problem that integrates task and motion planning. This results in dynamically feasible trajectories that satisfy safety requirements, including obstacle avoidance and mutual safety distances, while ensuring mission fulfillment.

    \item The proposed~\ac{STL} optimization problem (Section~\ref{sec:problemFormulation}) provides global optimal solutions. However, its nonlinear non-convex max-min nature presents challenges for solution methods~\cite{Bertsekas2012Book, CaiMacroeconomicDynamics2017}. To address this, we propose a hierarchical approach where an~\ac{MILP} approach (Section~\ref{sec:MILP}) generates the initial guess. This~\ac{MILP} approach works with simplified constraints, neglecting obstacle avoidance, safety requirements, vehicle dynamics, and time mission specifications. It outputs the sequence of target regions and corresponding refilling operations for each~\ac{UAV}. Motion primitives~\cite{Silano2021RAL} are then utilized to approximate~\ac{UAV} dynamics and generate feasible trajectories (position, velocity, and acceleration) between each pair of points. These trajectories serve as the initial guess for the global~\ac{STL} planner, aiding in its convergence towards solutions that meet all requirements.

    \item An event-based replanning strategy (Section~\ref{sec:replanner}) is introduced to handle unforeseen~\ac{UAV} failures, making the method applicable to scenarios with dynamic conditions. Additionally, an extension to the proposed planner (Section~\ref{sec:problemFormulation}) is presented in the form of an energy-aware planner (Section~\ref{sec:energyAwarePlanner}). This enhanced planner includes an energy minimization term to implicitly extend the endurance of the multi-rotors during the mission.

    \item Numerical simulations in MATLAB (Sections~\ref{sec:Validation}) assess the overall performance of the method in terms of mission specification fulfillment and the effectiveness of the novel initialization procedure for the~\ac{STL} optimization. Furthermore, Gazebo simulations and field experiments (Section~\ref{sec:fieldExperiments}) conducted in a mock-up setting demonstrate the validity and feasibility of the method for real scenarios. The Discussion section (Section~\ref{sec:discussion}) provides a critical analysis of the paper's findings, exploring their implications, limitations, and potential future extensions. Conclusions are discussed in Section~\ref{sec:conclusions}.
    
\end{itemize}



\section{Problem Description}
\label{sec:problemDescription}

\begin{figure}[tb]
    \centering
    \adjincludegraphics[width=\columnwidth]{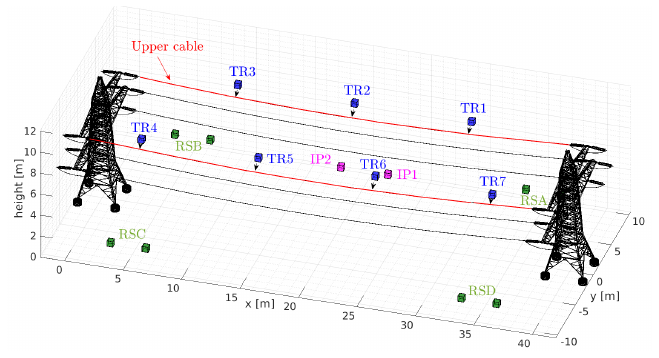}
    \caption{Illustration of a bird diverter installation scenario. Blue regions represent 
    \acfp{TR}, while the \acfp{IP} of \acp{UAV} and \acfp{RS} are depicted in magenta and 
    green, respectively. The red line denotes the upper cables connecting the power towers.}
    \label{fig:scenario}
\end{figure}

This paper focuses on the problem of installing bird diverters on power lines using a team of $\delta$~\acp{UAV}. 
The work presented here forms a part of the AERIAL-CORE European project\footnote{\url{https://aerial-core.eu}\label{ACproject}}. Figure~\ref{fig:scenario} illustrates the scenario setup. The objective is to visit specific \textit{target regions} along the upper cables between consecutive towers for installation purposes~\cite{Castano2021AppSci}. These regions serve as simplified representations of 3D spaces where diverters can potentially be installed. Once a~\ac{UAV} reaches a region, it is assumed that an onboard low-level controller handles the installation operation, e.g.,~\cite{Armengol2021AIRPHARO, Suarez2021AIRPHARO, CacaceDrones2023}. The~\acp{UAV} are assumed to be multi-rotors, specifically quadrotors, with limited velocity, acceleration, and payload capacity. Each~\ac{UAV} may have a different payload capacity, while the velocity and acceleration constraints remain the same for all~\acp{UAV}. Ground-based refilling stations are positioned along the power line, providing safe locations where the multi-rotors can load new diverters within a known and bounded time interval to resume their operations. The primary objective is to plan trajectories for the multi-rotors, ensuring that the mission is completed within the specified maximum time. This planning process must account for the vehicle dynamics and capacity constraints, as well as safety requirements such as obstacle avoidance and maintaining a safe distance between~\acp{UAV}. A map of the environment, which includes a polyhedral representation of obstacles such as power towers and cables, is assumed to be available prior to the installation mission.



\section{Preliminaries}
\label{sec:preliminaries}

Let us consider a discrete-time dynamical model of a multi-rotor system $x_{k+1}=f(x_k, u_k)$, where $x_{k+1}, x_k \in \mathcal{X} \in \mathbb{R}^n$ represent the next and current states of the system, respectively, and $u_k \in \mathcal{U} \in \mathbb{R}^m$ is the control input. Let us also assume that $f \colon \mathcal{X} \times \mathcal{U} \rightarrow \mathcal{X}$ is differentiable in both arguments and locally Lipschitz. Therefore, given an initial state $x_0 \in \mathcal{X}_0 \in \mathbb{R}^n$ and a time vector $\mathbf{t} = (t_0, \dots, t_N)^\top \in \mathbb{R}^{N+1}$, with $N \in \mathbb{N}_{>0}$ being the number of samples that describe the evolution of the system with the sampling period $T_s \in \mathbb{R}_{>0}$, we can define the finite control input sequence $\mathbf{u} = (u_0, \dots, u_{N-1})^\top \in \mathbb{R}^N$ as the input to provide the system to obtain the unique sequence of states $\mathbf{x} = (x_0, \dots, x_N)^\top \in \mathbb{R}^{N+1}$. 

Hence, we can define the state sequence $\mathbf{x}$ and the control input sequence $\mathbf{u}$ for the $d$-th multi-rotor, with $d \in \mathbb{N}_{> 0}$, as ${^d}\mathbf{x}=({^d}\mathbf{p}^{(1)}, {^d}\mathbf{v}^{(1)}, {^d}\mathbf{p}^{(2)}, {^d}\mathbf{v}^{(2)}, {^d}\mathbf{p}^{(3)}, {^d}\mathbf{v}^{(3)})^\top$ and ${^d}\mathbf{u}=({^d}\mathbf{a}^{(1)}, {^d}\mathbf{a}^{(2)}, {^d}\mathbf{a}^{(3)})^\top$, where ${^d}\mathbf{p}^{(j)}$, ${^d}\mathbf{v}^{(j)}$, and ${^d}\mathbf{a}^{(j)}$, with $j=\{1,2,3\}$, represent the sequences of position, velocity, and acceleration of the vehicle along the $j$-axis of the world frame, respectively. The $k$-th elements of ${^d}\mathbf{p}^{(j)}$, ${^d}\mathbf{v}^{(j)}$, ${^d}\mathbf{a}^{(j)}$, and $\mathbf{t}$ are denoted as ${^d}p^{(j)}_k$, ${^d}v^{(j)}_k$, ${^d}a^{(j)}_k$, and $t_k$, respectively, where $k \in \mathbb{N}_{>0}$.

The multi-rotor system model $x_{k+1} = f(x_k, u_k)$ employed in this paper is based on the formulation presented in the authors' prior work~\cite{Silano2021RAL}. In this formulation, motion primitives are represented using splines, offering advantages such as generating trajectories with arbitrary maneuver durations and state constraints. The computational burden is manageable, even on standard laptops, handling millions of motion primitives per second. Splines allow independent approximation of the multi-rotor's translational dynamics along each $j$-axis~\cite{MuellerTRO2015}. Here, we simply refer to them as $({^d}p_{k+1}^{(j)}, {^d}v_{k+1}^{(j)}, {^d}a_{k+1}^{(j)})^\top = {^d}\mathbf{S}^{(j)}({^d}p_{k}^{(j)}, {^d}v_{k}^{(j)}, {^d}a_{k}^{(j)})$. Throughout the following discussions, a label will be assigned to indicate the specific drone associated with the dynamic system, while the vector stack containing all drone variables will not include any explicit labels.



\subsection{Signal temporal logic}
\label{sec:SignalTemporalLogic}

\displayDefinitions[SignalTemporalLogic]{\acl{STL}}:~\ac{STL} was introduced for the first time in~\cite{maler2004FTMATFTS} to monitor the behavior of real-valued signals.~\ac{STL} provides a concise and unambiguous representation of complex system behaviors by encoding all mission specifications within a single formula $\varphi$~\cite{maler2004FTMATFTS}. For instance, it can express requirements such as ``at least two vehicles need to survey regions A and B, one must visit region C within the time interval $[t_1, t_2]$, while all vehicles must comply with safety requirements''. A comprehensive explanation of the~\ac{STL} syntax and semantics can be found in~\cite{maler2004FTMATFTS, donze2010ICFMATS}, but is omitted here for brevity. In short, an~\ac{STL} formula $\varphi$ is defined based on a set of predicates $p_i$, with $i \in \mathbb{N}_0$, which are atomic prepositions representing simple operations such as belonging to a region or comparisons of real values. These predicates can be combined using Boolean operators such as \textit{negation} ($\neg$), \textit{conjunction} ($\wedge$), \textit{disjunction} ($\vee$), and \textit{implication} ($\hspace{-0.45em} \implies \hspace{-0.45em}$), as well as temporal operators including \textit{eventually} ($\lozenge$), \textit{always} ($\square$), \textit{until} ($\mathcal{U}$), and \textit{next} ($\bigcirc$). The resulting~\ac{STL} formula $\varphi$ is considered valid if the expression evaluates to true ($\top$), and invalid ($\bot$) otherwise. For instance, informally, the expression $\varphi_1 \mathcal{U}_I \varphi_2$ implies that predicate $\varphi_2$ must hold at some point within the time interval $I$, and until then, predicate $\varphi_1$ must remain continuously satisfied. 



\subsection{Robust signal temporal logic}
\label{sec:robustSignalTemporalLogic}

\displayDefinitions[STLRobustness]{\ac{STL} Robustness}: The presence of system uncertainties, 
a dynamic environment, and unforeseen events can affect the satisfaction of an~\ac{STL} 
formula $\varphi$. To account for a margin of maneuverability in meeting $\varphi$, measuring 
how well (poorly) a given specification is satisfied, the concept of \textit{robust semantics} 
for~\ac{STL} formulae has been proposed~\cite{donze2010ICFMATS, maler2004FTMATFTS, 
Fainekos2009TCS}. This \textit{robustness}, denoted as $\rho$, is a quantitative metric that 
guides the optimization problem derived towards the best feasible solution to achieve mission 
satisfaction. This value can be formally described by using the following formulae in a 
recursive manner:\vspace{-1em}
\begin{equation*}
    \begin{array}{rll}
    \rho_{p_i} (\mathbf{x}, t_k) & = & \mu_i (\mathbf{x}, t_k), \\ 
    \rho_{\neg \varphi} (\mathbf{x}, t_k) & = & - \rho_\varphi (\mathbf{x}, t_k), 
    \\
    \rho_{\varphi_1 \wedge \varphi_2} (\mathbf{x}, t_k) & = & \min \left(\rho_{\varphi_1} (\mathbf{x}, t_k), \rho_{\varphi_2} (\mathbf{x}, t_k) \right), \\
    \rho_{\varphi_1 \vee \varphi_2} (\mathbf{x}, t_k) & = & \max \left(\rho_{\varphi_1} (\mathbf{x}, t_k), \rho_{\varphi_2} (\mathbf{x}, t_k) \right), \\
    \rho_{\square_I \varphi} (\mathbf{x}, t_k) & = & \min\limits_{t_k^\prime \in [t_k + I]} \rho_\varphi (\mathbf{x}, t_k^\prime), \\
    \rho_{\lozenge_I \varphi} (\mathbf{x}, t_k) & = & \max\limits_{t_k^\prime \in [t_k + I]} \rho_\varphi (\mathbf{x}, t_k^\prime), \\
    %
    %
    %
    \rho_{\bigcirc_I \varphi} (\mathbf{x}, t_k) & = & \rho_\varphi (\mathbf{x}, t_k^\prime), \text{with} \; t_k^\prime \in [t_k+I], \\
    \rho_{\varphi_1 \mathcal{U}_I \varphi_2} (\mathbf{x}, t_k) & = & \max\limits_{t_k^\prime \in [t_k + I]} \Bigl( \min \left( \rho_{\varphi_2} (\mathbf{x}, t_k^\prime) \right), \\ & & \hfill \min\limits_{ t_k^{\prime\prime} \in [t_k, t_k^\prime] } \left( \rho_{\varphi_1} (\mathbf{x}, t_k^{\prime \prime} \right)  \Bigr),
    \end{array}
\end{equation*}
where $t_k + I$ represents the Minkowski sum of the scalar $t_k$ and the time interval $I$. The formulae described above consist of a set of predicates, denoted as $p_i$, along with their corresponding real-valued functions $\mu_i(\mathbf{x}, t_k)$. These predicates are considered true if their robustness value is greater than zero, and false otherwise. 

The entire formula operates as a logical expression, where it evaluates to false if at least one predicate is false. The evaluation follows the application of logical and temporal operators (such as \textit{always}, \textit{eventually}, \textit{conjunction}, etc.) from the innermost part to the outermost part of the formula. For instance, an example could involve being inside a target region or outside an obstacle region, with the regions being defined by a specific number of predicates. Further details can be found in~\cite{Souza2007JSTTT, maler2004FTMATFTS, donze2010ICFMATS}. In this case, we say that $\mathbf{x}$ satisfies the~\ac{STL} formula $\varphi$ at time $t_k$ if $\rho_\varphi(\mathbf{x}, t_k) > 0$ (in short, denoted as $\mathbf{x}(t_k)\models\varphi$), and $\mathbf{x}$ violates $\varphi$ if $\rho_\varphi(\mathbf{x}, t_k) \leq 0$. To simplify notation, we will use $\rho_\varphi(\mathbf{x})$ instead of $\rho_\varphi(\mathbf{x},0)$ when $t_k = 0$.

Thus, we can compute the control inputs $\mathbf{u}$ that maximize the robustness $\rho_\varphi(\mathbf{x})$ over a set of finite state and input sequences $\mathbf{x}$ and $\mathbf{u}$, respectively. An optimal sequence $\mathbf{u}^\star$ is considered valid if $\rho_\varphi (\mathbf{x}^\star)$ is positive, where $\mathbf{x}^\star$ and $\mathbf{u}^\star$ obey the dynamical system. A higher value of $\rho_\varphi (\mathbf{x}^\star)$ indicates a more robust behavior of the system against disturbances, enabling the system to tolerate higher levels of disruption without violating the~\ac{STL} specification.

\displayDefinitions[SmoothApproximation]{Smooth Approximation}~\cite{Gilpin2021LCSS}: Let us 
consider $\lambda \in \mathbb{R}_{>0}$ as a tunable parameter. The smooth approximation of the 
$\min$ and $\max$ operators with $\beta$-predicate arguments is:\vspace{-1em}
\begin{equation*}
    \begin{split}
    &\max(\rho_{\varphi_1}, \dots, \rho_{\varphi_\beta}) \approx \frac{ \sum_{i=1}^\beta  \rho_{\varphi_i} e^{\lambda \rho_{\varphi_i}} }{ \sum_{i=1}^\beta  e^{\lambda \rho_{\varphi_i}} }, 
    \\
    &\min(\rho_{\varphi_1}, \dots, \rho_{\varphi_\beta}) \approx -\frac{1}{\lambda} \log \left( \sum_{i=1}^\beta e^{-\lambda \rho_{\varphi_i}} \right). 
    \end{split}
\end{equation*}\vspace{-0.5em}

In this paper, we introduce an improved approximation approach compared to our previous work~\cite{Silano2021RAL}. This new method possesses the properties of being \textit{asymptotically complete} and \textit{smooth everywhere}, similar to the well-known \ac{LSE} approximation~\cite{donze2010ICFMATS}. Furthermore, it guarantees \textit{soundness} by ensuring that an optimal sequence $\mathbf{u}^\star$ with strictly positive smooth robustness ($\tilde{\rho}_\varphi(\mathbf{x}) > 0$) satisfies the specification $\varphi$, while an optimal sequence $\mathbf{u}^\star$ with negative smooth robustness ($\tilde{\rho}_\varphi(\mathbf{x}) \leq 0$) violates it. The property of \textit{asymptotical completeness} indicates that as the parameter $\lambda$ tends towards infinity, the approximation of the final smooth robustness formula $\tilde{\rho}_\varphi(\mathbf{x})$ can be arbitrarily close to the true robustness $\rho_\varphi(\mathbf{x})$. By achieving \textit{smoothness everywhere}, the approximation is infinitely differentiable, enabling the utilization of gradient-based optimization algorithms to find solutions to the problem at hand~\cite{Gilpin2021LCSS}. 

\displayDefinitions[STLMotionPlanner]{STL Motion Planner}~\cite{Silano2021RAL}: Starting from 
mission specifications encoded as~\ac{STL} formula $\varphi$, and replacing its robustness 
$\rho_\varphi(\mathbf{x})$ with the smooth approximation\footnote{The computation of 
$\rho_\varphi(\mathbf{x})$ involves non-differentiable functions such as $\min$ and $\max$. To 
address the computational complexity associated with these non-differentiable functions, it is 
beneficial to employ a smooth approximation, denoted as $\tilde{\rho}_\varphi(\mathbf{x})$, of 
the robustness function $\rho_\varphi(\mathbf{x})$. This smooth approximation provides a more 
tractable and computationally efficient alternative.} $\tilde{\rho}_\varphi(\mathbf{x})$ 
(\refDefinitions[SmoothApproximation]), the generation of multi-rotor trajectories can be cast 
as an optimization problem:\vspace{-1em}
\begin{equation}\label{eq:optimizationProblemMotionPrimitives}
    \begin{split}
    &\maximize_{\mathbf{p}^{(j)}, \, \mathbf{v}^{(j)}, \, \mathbf{a}^{(j)} \atop d \in \mathcal{D}} \;\;
    {\tilde{\rho}_\varphi (\mathbf{p}^{(j)}, \mathbf{v}^{(j)} )} \\
    %
    %
    &\qquad \text{s.t.}~\quad\; {^d}\underline{v}^{(j)} \leq {^d}v^{(j)}_k \leq {^d}\bar{v}^{(j)},  \\
    &\,\;\;\, \qquad \quad\;\;\, {^d}\underline{a}^{(j)} \leq {^d}a^{(j)}_k \leq {^d}\bar{a}^{(j)}, \; \\
    &\,\;\;\, \qquad \quad\;\;\, \tilde{\rho}_\varphi (\mathbf{p}^{(j)}, \mathbf{v}^{(j)} ) \geq \varepsilon, \\
    &\,\;\;\, \qquad \quad\;\;\, {^d}\mathbf{S}^{(j)}, \forall k=\{0,1, \dots, N-1\},
    \end{split},
\end{equation}
where ${^d}\bar{v}^{(j)}$ and ${^d}\bar{a}^{(j)}$ represent the desired maximum values for velocity and acceleration, respectively, of drone $d$ along each $j$-axis of the world frame. The set of drones is denoted as $\mathcal{D}$. The lower bound on robustness, $\tilde{\rho}_\varphi (\mathbf{p}^{(j)}, \mathbf{v}^{(j)}) \geq \varepsilon$, provides a safety margin for satisfying the~\ac{STL} formula $\varphi$ in the presence of disturbances. As demonstrated in~\cite{Pant2017CCTA}, disturbances with magnitudes smaller than $\varepsilon$ do not lead to a violation of the formula. The value of $\varepsilon$ can be computed such that $\lvert \rho_\varphi(\mathbf{x}) - \tilde{\rho}_\varphi(\mathbf{x}) \rvert \leq \varepsilon$. Finally, ${^d}\mathbf{S}^{(j)}$ refers to the motion primitives employed to describe the motion of drone $d$ along each $j$-axis, as previously mentioned in the Preliminaries.



\section{Problem Solution}
\label{sec:problemFormulation}

In this section, we utilize the~\ac{STL} framework presented in Section~\ref{sec:preliminaries} to formulate the optimization problem outlined in Section~\ref{sec:problemDescription}. This formulation results in a nonlinear non-convex max-min problem, which we represent as a~\ac{NLP} formulation and solve using dynamic programming techniques (Section~\ref{sec:motionPlanner}). Solving this type of nonlinear problem within a reasonable time frame is challenging due to the propensity of solvers to converge to local optima~\cite{Bertsekas2012Book, CaiMacroeconomicDynamics2017}. To address this issue, we compute an initial guess using a simplified~\ac{MILP} formulation in Section~\ref{sec:MILP}, which does not consider obstacle avoidance, safety requirements, vehicle dynamics, and time mission specifications. This simplification facilitates the search for a global solution. Additionally, our proposed framework includes a mechanism for mission replanning in the event of~\ac{UAV} failures (Section~\ref{sec:replanner}). Finally, we enhance the original motion planner by minimizing energy consumption, leading to implicit savings in multi-rotor flight time (Section~\ref{sec:energyAwarePlanner}). 



\subsection{STL motion planner}
\label{sec:motionPlanner}

In this section, we extract the mission specifications for the problem discussed in Section~\ref{sec:problemDescription} to obtain the corresponding~\ac{STL} formula, $\varphi$. The successful installation of bird diverters requires a collaborative team of~\acp{UAV} that can effectively respond to unforeseen events, cover large-scale scenario and minimize operation time. The mission specifications can be categorized into two types: safety requirements and task requirements. The \textit{safety requirements} ensure the overall safety of the operation throughout the entire mission time $T_N$. These requirements include constraints such as the~\acp{UAV} staying within the designated workspace (${^d}\varphi_\mathrm{ws}$), avoiding collisions with obstacles in the environment (${^d}\varphi_\mathrm{obs}$), and maintaining a safe distance from other~\acp{UAV} (${^d}\varphi_\mathrm{dis}$). On the other hand, the \textit{task requirements} focus on achieving specific tasks at predefined time intervals during the mission. These requirements involve ensuring that all target regions are visited by at least one~\ac{UAV} (${^d}\varphi_\mathrm{tr}$) and that each~\ac{UAV} remains in a target region for the designated installation time $T_\mathrm{ins}$. Considering the limited payload capacity of the~\acp{UAV}, they are required to visit a refilling station and remain there for a refilling time $T_\mathrm{rs}$ once they run out of onboard diverters (${^d}\varphi_\mathrm{rs}$). Notably, we do not explicitly specify the number or sequence of target regions that a single drone must visit. Rather, we grant the framework the flexibility to determine the most convenient number and sequence of targets for each vehicle, while still allowing for the possibility of visiting all targets if needed. Finally, after completing their installation operations, each~\ac{UAV} should fly to the closest refilling station (${^d}\varphi_\mathrm{hm}$). By incorporating all these mission specifications, the~\ac{STL} formula can be formulated as follows:
\begin{equation}\label{eq:longRangeInspection}
    \resizebox{1\hsize}{!}{$%
    \begin{split}
    \varphi =&  \bigwedge_{d\in\mathcal{D}}\square_{[0,T_N]} ( {^d}\varphi_{\mathrm{ws}} \wedge {^d}\varphi_{\mathrm{obs}} \wedge {^d}\varphi_{\mathrm{dis}} ) \,
    \wedge \\
    & \bigwedge_{q=1}^\mathrm{tr}\lozenge_{[0,T_N-T_{\mathrm{ins}}]} \bigvee_{d\in\mathcal{D}}\square_{[0,T_{\mathrm{ins}}]} \left({^d}c(t_k) > 0 \right) {^d}\varphi_{\mathrm{tr,q}} \, \wedge \\
    & \bigwedge_{d\in\mathcal{D}} \lozenge_{[0,T_N-T_{\mathrm{rs}}]} \bigvee_{q=1}^\mathrm{rs}\square_{[0,T_{\mathrm{rs}}]}  \left({^d}c(t_k) = 0 \hspace{-0.3em} \implies \hspace{-0.3em} {^d}\mathbf{p}(t_k)\models {^d}\varphi_{\mathrm{rs,q}}\right) \wedge \\
    & \bigwedge_{d\in\mathcal{D}} \square_{[1,T_N-1]}\left({^d}\mathbf{p}(t_k)\models {^d}\varphi_{\mathrm{hm}} \hspace{-0.3em} \implies \hspace{-0.3em} {^d}\mathbf{p}(t_k+1)\models {^d}\varphi_{\mathrm{hm}} \right).
   \end{split}
    $}%
\end{equation}

The~\ac{STL} formula $\varphi$ consists of six specifications (${^d}\varphi_\mathrm{ws}$, ${^d}\varphi_\mathrm{obs}$, ${^d}\varphi_\mathrm{dis}$, ${^d}\varphi_\mathrm{tr}$, ${^d}\varphi_\mathrm{rs}$, and ${^d}\varphi_\mathrm{hm}$) and three time intervals ($T_N$, $T_\mathrm{ins}$, and $T_\mathrm{rs}$). The following equations describe each of these specifications:
\begin{subequations}\label{eq:STLcomponents}
    \begin{align}
    \textstyle{{^d}\varphi_\mathrm{ws}} &= \textstyle{\bigwedge_{j=1}^3 \mathbf{p}^{(j)} \in (\underline{p}^{(j)}_\mathrm{ws}, \bar{p}^{(j)}_\mathrm{ws})}, \label{subeq:belongWorkspace} \\
    \textstyle{{^d}\varphi_\mathrm{obs}} &= \textstyle{ \bigwedge_{j=1}^3\bigwedge_{q=1}^{\mathrm{obs}} \mathbf{p}^{(j)} \hspace{-0.25em} \not\in (\underline{p}_{\mathrm{obs,q}}^{(j)}, \bar{p}_{\mathrm{obs,q}}^{(j)})}, \label{subeq:avoidObostacles} \\
    \textstyle{{^d}\varphi_\mathrm{dis}} &=  \textstyle{\bigwedge_{ \{d,m\} \in \mathcal{D}, d \neq m } \hspace{0.2em} \lVert {^d}\mathbf{p} - {^m}\mathbf{p} \rVert \geq \Gamma_\mathrm{dis}}, \label{subeq:keepDistance} \\
    \textstyle{{^d}\varphi_\mathrm{hm}} &= \textstyle{\bigwedge_{j=1}^3 \mathbf{p}^{(j)} \hspace{-0.25em} \in (\underline{p}^{(j)}_\mathrm{hm}, \bar{p}^{(j)}_\mathrm{hm})}, \label{subeq:backHome} \\
    \textstyle{{^d}\varphi_{\mathrm{tr,q}}} &=  \textstyle{\bigwedge_{j=1}^3  {^d}\mathbf{p}^{(j)} \hspace{-0.25em} \in (\underline{p}^{(j)}_{\mathrm{tr,q}}, \bar{p}^{(j)}_{\mathrm{tr,q}}) \, \wedge} \nonumber \\
    &  \textstyle{\wedge \bigcirc_{T_\mathrm{ins}} \left( {^d}c(t_k) = {^d}c(t_k-1) - 1 \right)}, \label{subeq:visitTargets} \\
    \textstyle{{^{d}}\varphi_\mathrm{rs,q}} &= \textstyle{\bigwedge_{j=1}^3 \mathbf{p}^{(j)} \hspace{-0.25em} \in (\underline{p}_\mathrm{rs,q}^{(j)}, \bar{p}_\mathrm{rs,q}^{(j)}) \, \wedge} \nonumber \\
    &  \textstyle{\wedge \bigcirc_{T_\mathrm{rs}} \left( {^d}c(t_k) = {^d}\bar{c} \right)}, \label{subeq:refill}
    \end{align}
\end{subequations} 
where~\eqref{subeq:belongWorkspace} constrains the position of each~\ac{UAV} to remain within the designated workspace, with $\underline{p}^{(j)}_\mathrm{ws}$ and $\bar{p}^{(j)}_\mathrm{ws}$ representing the working space limits. Obstacle avoidance and mission completion operation are defined in~\eqref{subeq:avoidObostacles} and~\eqref{subeq:backHome}, respectively. Rectangular regions with vertices denoted by $\underline{p}_\mathrm{obs,q}^{(j)}$, $\underline{p}_\mathrm{hm}^{(j)}$, $\bar{p}_\mathrm{obs,q}^{(j)}$, and $\bar{p}_\mathrm{hm}^{(j)}$ define obstacle and drone's final position (closest refilling station) areas. The safety distance requirement is encoded in~\eqref{subeq:keepDistance}, where $\Gamma_\mathrm{dis} \in \mathbb{R}_{>0}$ represents the threshold value for the mutual distance between~\acp{UAV}. Finally, visiting target areas and refilling stations are described by~\eqref{subeq:visitTargets} and~\eqref{subeq:refill}, respectively. The vertices $\underline{p}_\mathrm{tr,q}^{(j)}$, $\underline{p}_\mathrm{rs,q}^{(j)}$, $\bar{p}_\mathrm{tr,q}^{(j)}$, and $\bar{p}_\mathrm{rs,q}^{(j)}$ identify the target and refilling areas. In~\eqref{subeq:visitTargets} and~\eqref{subeq:refill}, the payload capacity ${^d}c(t_k) \in \mathbb{N}_0$ is used to represent the number of remaining and loaded diverters for each~\ac{UAV}, respectively. The payload capacity is constrained to the interval ${^d}c(t_k) \in [0, {^d}\bar{c}]$, where ${^d}\bar{c}$ corresponds to the nominal payload capacity. The \textit{always} operators ($\square$) guarantee the fulfillment of the time requirements $T_\mathrm{ins}$ and $T_\mathrm{rs}$, which correspond to the durations of the installation and refilling operations, respectively. On the other hand, the \textit{next} operators ($\bigcirc$) ensure that changes in the payload capacity will occur only following those time intervals Additionally, the \textit{eventually} operators ($\lozenge$) ensure that the mission requirements will ultimately be fulfilled within the designated mission time $T_N$.

Using the specifications described in~\eqref{eq:longRangeInspection}, the optimization problem defined in~\refDefinitions[STLMotionPlanner] is formulated to compute feasible trajectories that maximize the smooth robustness $\tilde{\rho}_\varphi(\mathbf{x})$~\ac{wrt} the given mission specifications $\varphi$. To achieve this, it is necessary to compute the robustness score for each predicate. The~\ac{STL} formula~\eqref{eq:longRangeInspection} comprises two types of predicates. The first type evaluates whether the~\ac{UAV} position belongs or does not belong to a specific region, as depicted in~\eqref{subeq:belongWorkspace},~\eqref{subeq:avoidObostacles},~\eqref{subeq:backHome},~\eqref{subeq:visitTargets}, and~\eqref{subeq:refill}. The second type evaluates the distance between~\acp{UAV}, as represented by the safety requirement in~\eqref{subeq:keepDistance}. The robustness values are quantified based on the Euclidean distance. For predicates belonging to the first group, a positive robustness indicates that the~\ac{UAV} lies within the designated region. The robustness increases as the minimum Euclidean distance to the boundaries of the region along the trajectory becomes larger. However, in the case of~\eqref{subeq:avoidObostacles}, the inverse applies, where being within the obstacle region corresponds to a negative robustness. In the safety distance predicate~\eqref{subeq:keepDistance}, the robustness is positive when the distance between~\acp{UAV} exceeds the threshold $\Gamma_\mathrm{dis}$. The robustness value increases as the minimum Euclidean distance between~\acp{UAV} along the trajectory becomes larger. To illustrate this concept further, we can consider the ${^d}\varphi_\mathrm{ws}$ predicate~\eqref{subeq:belongWorkspace} as an example:
\begin{equation}\label{eq:predicatesEqu}
    \resizebox{1\hsize}{!}{$%
    {^d}\rho_{\varphi_{\mathrm{ws}}} =  \min\limits_{\mathbf{p}^{(j)}} \left (\min ( { {^d}\rho_{\bar{\varphi}^{(1)}} }, { {^d}\rho_{\underline{\varphi}^{(1)}} }, { {^d}\rho_{\bar{\varphi}^{(2)}} }, { {^d}\rho_{\underline{\varphi}^{(2)}} }, { {^d}\rho_{\bar{\varphi}^{(3)}} }, { {^d}\rho_{\underline{\varphi}^{(3)}} } ) \right ),
    $}%
\end{equation}
with
\begin{align*}
    { {^d}\rho_{\bar{\varphi}^{(j)}} } &= \bar{p}_\mathrm{ws}^{(j)} - {^d}p_k^{(j)},
    \quad
    { {^d}\rho_{\underline{\varphi}^{(j)}} } = {^d}p_k^{(j)} - \underline{p}_\mathrm{ws}^{(j)}. 
\end{align*}

Similarly, the robustness score of the non-belonging predicate ${^d}\varphi_\mathrm{obs}$~\eqref{subeq:avoidObostacles} can be computed by taking the inverse of each minimum distance for each sample along the trajectory. In other words, we can express this as:
\begin{equation}\label{eq:predicatesObsExplained}
    \resizebox{1\hsize}{!}{$%
    {^d}\rho_{\varphi_{\mathrm{obs}}} =  \min\limits_{\mathbf{p}^{(j)}} \left (-\min ( { {^d}\rho_{\bar{\varphi}^{(1)}} }, { {^d}\rho_{\underline{\varphi}^{(1)}} }, { {^d}\rho_{\bar{\varphi}^{(2)}} }, { {^d}\rho_{\underline{\varphi}^{(2)}} }, { {^d}\rho_{\bar{\varphi}^{(3)}} }, { {^d}\rho_{\underline{\varphi}^{(3)}} } ) \right ),
    $}%
\end{equation}
where ${^d}\rho_{\bar{\varphi}^{(j)}}$ and ${^d}\rho_{\underline{\varphi}^{(j)}}$, with $j=\{1,2,3\}$, can be computed by replacing $\bar{p}^{(j)}_\mathrm{ws}$ and $\underline{p}^{(j)}_\mathrm{ws}$ in~\eqref{eq:predicatesEqu} with $\bar{p}^{(j)}_\mathrm{obs}$ and $\underline{p}^{(j)}_\mathrm{obs}$, respectively.
On the contrary, the robustness score of the safety requirement predicate ${^d}\varphi_\mathrm{dis}$~\eqref{subeq:keepDistance} can be expressed as:
\begin{equation}
    {^d}\rho_{\varphi_{\mathrm{dis}}} = \min\limits_{\mathbf{p}^{(j)}} \left( \min\limits_{ \{d,m\} \in \mathcal{D}, \, d \neq m} ( \lVert {^d}\mathbf{p} - {^m}\mathbf{p} \rVert - \Gamma_\mathrm{dis} ) \right),
\end{equation}
where the minimum distance between~\acp{UAV} is computed for each time step ($t_k$) in the trajectory, and then the $\min$ value of that vector of minimum distances.

It is important to emphasize that the interdependence of mission requirements necessitates the joint planning of all~\ac{UAV} trajectories through a unified optimization problem. 



\subsection{Initial guess}
\label{sec:MILP}

To ensure optimal solutions within a reasonable time and prevent the solver from becoming trapped, it is essential to have an appropriate initial guess for the~\ac{STL} motion planner. The key concept behind obtaining this initial guess is to simplify the original diverter installation problem by abstracting it into an optimization with fewer constraints. Specifically, the initial guess focuses on fulfilling mission requirements related to visiting target regions, considering refilling and mission completition operations (${^d}\varphi_\mathrm{tr}$, ${^d}\varphi_\mathrm{rs}$, and ${^d}\varphi_\mathrm{hm}$). It omits obstacle avoidance, workspace, and safety distance requirements (${^d}\varphi_\mathrm{obs}$, ${^d}\varphi_\mathrm{ws}$, and ${^d}\varphi_\mathrm{dis}$), and the mission time intervals ($T_N$, $T_\mathrm{ins}$, and $T_\mathrm{rs}$) since these constraints introduce complex nonlinearities and motion discontinuities in the problem. To generate the initial guess, a graph-based representation is employed, establishing connections between target regions and refilling stations. This formulation is modeled as a variant of the Vehicle Routing Problem (\ac{VRP}) using~\acl{MILP}. Assigning target regions to vehicles and providing navigation sequences for each~\ac{UAV} are the objectives of the~\ac{MILP} solution. 

\begin{figure}[tb]
	\centering
	\scalebox{1}{
		\adjincludegraphics[width=\columnwidth]{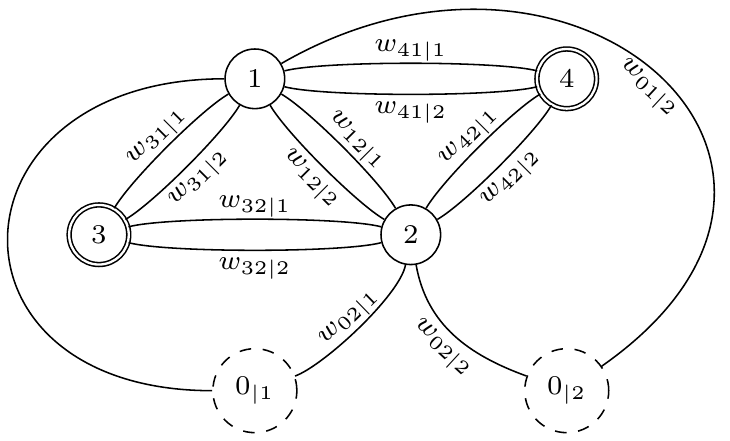}
	}
	\caption{Instance of the graph $G$ assuming two target regions (solid round nodes), two 
		refilling stations (double round nodes), and two vehicles. The paths of the~\acp{UAV} 
		are 
		represented by arcs with corresponding weights $w_{ij \| d}$. The depots are depicted 
		as 
		dashed round nodes.}
	\label{fig:undirectedMultigraph}
\end{figure}

Figure~\ref{fig:undirectedMultigraph} illustrates the graph used in our approach, which is an undirected weighted multigraph denoted by the tuple $G = (\mathcal{V}, \mathcal{E}, \mathcal{W}, \mathcal{D}, \mathcal{C})$. The set of vertices, $\mathcal{V}$, consists of locations that encompass target regions, refilling stations, and depots where each~\ac{UAV} is initially positioned at time $t_0$. Specifically, $\mathcal{V} = \mathcal{T} \cup \mathcal{R} \cup \mathcal{O}$, with $\lvert \mathcal{T} \rvert = \tau$ representing the target regions, $\lvert \mathcal{R} \rvert = r$ representing the refilling stations, and $\lvert \mathcal{O} \rvert = \delta$ including the depots. The set of edges, $\mathcal{E}$, and their corresponding weights, $\mathcal{W}$, define the connectivity and distances between the vertices, respectively. Additionally, as said before, $\mathcal{D}$ represents the set of~\acp{UAV}, with $\lvert \mathcal{D} \rvert = \delta$, and $\mathcal{C} = \{{^1}\bar{c}, \dots, {^\delta}\bar{c}\}$ denotes their respective maximum payload capacities.

Regarding the graph connectivity, all vertices in $\mathcal{T}$ are fully connected to each 
other and connected to every vertex in the set $\mathcal{R} \cup \mathcal{O}$. Each connection 
involves one edge per~\ac{UAV}, totaling $\delta$ edges. More formally, let $e_{ij \| d} \in 
\mathcal{E}$ be the edge that connects vertices $i$ and $j$, with $\{i,j\} \in \mathcal{V}$ 
and $i \neq j$, using~\ac{UAV} $d \in \mathcal{D}$, where $d=\{1, \dots, \delta \}$. The 
weight associated with $e_{ij \| d}$ is denoted as $w_{ij \| d} \in \mathcal{W}$. Considering 
the homogeneous dynamic constraints for~\acp{UAV}, such as maximum velocities 
${^d}\bar{v}^{(j)}$ and maximum accelerations ${^d}\bar{a}^{(j)}$, and assuming similar flight 
durations for each~\ac{UAV}, the mission operation constraint can be expressed as a maximum 
distance traveled by each individual~\ac{UAV}. As a result, we model the edge weights using 
Euclidean distances, resulting in $w_{ij \| d} = w_{ji \| d}$.


For each edge $e_{ij \| d} \in \mathcal{E}$, we introduce an integer variable $z_{ij \| d} \in 
\mathbb{Z}_ 0$. This variable represents the number of times the corresponding edge is 
selected in the solution of the~\ac{MILP}. Thus, for a given~\ac{UAV} $d \in \mathcal{D}$, we 
have $z_{ij \| d} \in \{0,1\}$ for all $\{i,j\} \in \{ \mathcal{T},\mathcal{O} \}$, and $z_{ij 
\| d} \in \{0,1,2\}$ if $i \in \mathcal{R}$ and $j \in \mathcal{T}$. The former ensures that 
the edge between two target regions is never covered twice, indicating that the~\ac{UAV} 
starts from a depot and never returns to it. The latter allows for round trips between 
refilling stations and target regions, in case there are no other diverters to be installed. 
Specifically, $z_{ij \| d} = 2$ denotes a return trip between a refilling station $i$ and a 
target region $j$ of drone $d$. When the subscript of the variable $z_{ij \| d}$ is $0$, it 
indicates that the corresponding vertex $i$ or $j$ is a depot, depending on the order. %
Utilizing these variables, we can formulate the~\ac{MILP} problem as follows: \vspace{-1em}
\begin{subequations}\label{eq:MILP}
    \begin{align}
        &\minimize_{z_{ij \| d}, y_{j \| d}}
        { \sum\limits_{ \{i,j\} \in \mathcal{V}, \, i \neq j, \, d \in \mathcal{D}} 
        \hspace{-2em} w_{ij \| d} \, z_{ij \| d} } \label{subeq:objectiveFunction} \\
        %
        &\;\;\;\;\;\; \text{s.t.} \;\, \hspace{-0.45cm} \sum\limits_{i \in \mathcal{V}, \, i 
        \neq j, \, d \in \mathcal{D}} \hspace{-0.975em} z_{ij \| d} = 2, \; \forall j \in 
        \mathcal{T}, \label{subeq:visitedOnce} \\ 
        &\quad \, \;\;\,\;\;\;\; \sum\limits_{ i \in \mathcal{V}, \, i \neq j} \quad 
        \hspace{-0.975em} z_{ij \| d} = 2y_{j \| d}, \; \forall j \in \mathcal{T}, \; \forall 
        d 
        \in \mathcal{D}, \label{subeq:visitedOneUAV} \\ 
        %
        &\qquad \;\,\;\;\;\;\; \sum\limits_{ i \in \mathcal{T} } \qquad \hspace{-1.23em} z_{0i 
        \| d} = 1, \; \forall d \in \mathcal{D}, \label{subeq:depotVisitedOnce} \\  
        %
        & \;\;\;\;\;\;\; \sum\limits_{\substack{ i \in \mathcal{T}, \, j \not\in \mathcal{T}, 
        \, d \in \mathcal{D}}} \hspace{-1em} z_{ij \| d} \geq 2 
        h\hspace{-0.2em}\left(\mathcal{T}\right). \label{subeq:capacityAndSubtours} 
        %
        %
        %
        %
     \end{align}
\end{subequations}

In the above formulae, the objective function~\eqref{subeq:objectiveFunction} quantifies the 
total distance covered by the multi-\ac{UAV} team. Constraints~\eqref{subeq:visitedOnce} 
and~\eqref{subeq:visitedOneUAV} ensure that each target region is visited exactly once. To 
achieve this, auxiliary integer variables $y_{j \| d} \in \{0,1\}$ are introduced, which 
ensure that if a~\ac{UAV} $d \in \mathcal{D}$ reaches target $j \in \mathcal{T}$, the 
same~\ac{UAV} must also leave the target. Constraint~\eqref{subeq:depotVisitedOnce} guarantees 
that each~\ac{UAV} starts the mission from its depot and does not return to it. 
Constraints~\eqref{subeq:capacityAndSubtours} serve two purposes: preventing tours that exceed 
the payload capacity of the~\acp{UAV} and ensuring that all tours connect to a refilling 
station, which is commonly known as the sub-tour elimination constraint~\cite{Miller1960, 
LaporteNRL2007}. The lower bound $h(\mathcal{T})$ represents the minimum number of~\acp{UAV} 
required to visit all target regions $\mathcal{T}$. In practice, $h(\mathcal{T})$ is often 
defined as $\lceil \sum_{i \in \mathcal{T}} c_i / \mathcal{C} \rceil$, where $c_i$ is the 
number of diverters at target $i$ and $\mathcal{C}$ is the payload capacity of the~\acp{UAV}. 
To address the computational challenges posed by the exponential number of constraints associated with $h(\mathcal{T})$~\cite{Laporte1986JOPRS, Yuan2020OPL, Achuthan1996EJOR}, an adaptive constraint incorporation strategy is employed. This approach dynamically adds constraints as violations occur during the optimization process~\cite{LaporteNRL2007}. By incorporating constraints only when necessary, based on violations identified during the solution process, the computational burden is alleviated. This adaptive constraint management strategy improves the computational efficiency of the solution by avoiding the exhaustive evaluation of all possible combinations and focusing on constraints that are most relevant or impactful. 

%
%

After solving the~\ac{MILP} problem, the optimal assignment of target regions and refilling 
stations for each~\ac{UAV} is obtained. To generate dynamically feasible trajectories, we 
employ multi-rotor motion primitives~\cite{Silano2021RAL}, as said in the Preliminaries 
(Section~\ref{sec:preliminaries}). The detailed computation of these trajectories is provided 
in~\cite{MuellerTRO2015} and is not presented here for brevity. In summary, the motion 
primitives are computed by assuming rest-to-rest motion between regions, where the~\ac{UAV} 
has zero velocity and acceleration. The minimum feasible time required to satisfy the desired 
maximum values of velocity ${^d}\bar{v}^{(j)}$ and acceleration ${^d}\bar{a}^{(j)}$ along 
the~\ac{UAV}'s trajectory is imposed. The trajectories also account for the installation time 
intervals $T_\mathrm{ins}$ and refilling time intervals $T_\mathrm{rs}$ during which 
the~\acp{UAV} remain stationary at the corresponding regions. 



\subsection{Event-triggered replanner}
\label{sec:replanner}

The presented motion planner enables the generation of feasible trajectories for a multi-\ac{UAV} team carrying out a bird diverter installation mission. However, in real-world scenarios,~\ac{UAV} failures can occur (e.g., due to a battery or technical fault), resulting in the need for an alternative~\ac{UAV} to take over the pending tasks to ensure the successful completion of the mission. To address this, an online event-based replanning procedure is employed.

Whenever a~\ac{UAV} failure event is detected, a new plan is computed online using the motion planner described in Sections~\ref{sec:motionPlanner} and~\ref{sec:MILP}. The trajectories of the~\acp{UAV} that are still operational are not recomputed since they can continue with their original plans. The replanning process focuses on generating a new trajectory for the \textit{backup}~\ac{UAV} that replaces the faulty one. Assuming the failure event occurs at time $t_\mathrm{fail} \in [t_0,T_N]$, the new trajectory for the backup~\ac{UAV} is generated within the time interval $[t_\mathrm{fail} + T_\mathrm{rep}, T_N]$. Here, $T_\mathrm{rep}$ represents the maximum expected computation time for replanning, which was estimated by running multiple instances of the~\ac{STL} optimization problem and varying factors such as the number and positions of target regions, as well as the time instance $t_\mathrm{fail}$ when the failure occurs.

During the replanning process, the motion planner utilizes the original trajectories of the non-faulty~\acp{UAV} within the time interval $[t_\mathrm{fail} + T_\mathrm{rep}, T_N]$ as input constraints for the safety distance requirement (${^d}\varphi_\mathrm{dis}$). As said before, these trajectories do not need to be recomputed. Additionally, only the pending target regions that were previously assigned to the faulty~\ac{UAV} are considered in the replanning procedure. However, the order in which these target regions are visited may change, as the backup~\ac{UAV} may start from a different position or with a different payload, potentially resulting in a suboptimal visitation sequence. 

It is important to note that the computational effort required for replanning is significantly 
lower than that for initial planning, as it involves only a single~\ac{UAV} and a reduced set 
of target regions (those that have not yet been visited). While exploring alternative 
re-planning strategies for the routes of non-faulty drones may potentially yield more optimal 
solutions in terms of robustness scores, it is essential to emphasize that our paper focuses 
on ensuring operational continuity, minimizing disruptions, and maintaining safety in 
hazardous scenarios. Prioritizing the better trajectories solution in these cases may lead to 
the drones halting their mission, resulting in delays and increased costs for the entire 
operation. Therefore, our approach offers a rapid and effective solution by replacing a faulty 
multi-rotor with a backup vehicle, ensuring seamless continuation of the mission. \vspace{-2em}



\subsection{Energy-aware planner}
\label{sec:energyAwarePlanner}

This section introduces an energy-aware extension of the motion planner presented in Section~\ref{sec:motionPlanner}, focusing on minimizing the energy consumption of the~\acp{UAV} during the mission. This results in implicit savings in multi-rotor flight time. The key concept is to determine the optimal cruising speed, considering the aerodynamics of the multi-rotors, to achieve minimal energy usage. The goal is to generate trajectories for the~\acp{UAV} that closely align with these optimal speeds.

The power required by a multi-rotor decreases as the low forward speed increases, while it 
increases significantly at high speeds due to parasitic losses. This combination of effects 
gives rise to an optimal forward speed that minimizes the energy consumption of the~\ac{UAV} 
and maximizes its flight time. This energetically optimal forward speed, denoted as $v^\star$, 
is described in~\cite[Ch.~5]{Leishman2006CUP}, and defined as: \vspace{-1em}

\begin{equation}\label{eq:velocityInftyStar}
v^\star = \sqrt{\frac{ m }{ 2 \varrho A }} \left( \frac{ 4 \kappa A }{ n_r f } \right)^{1/4} ,
\end{equation}
where $m \in \mathbb{R}$ represents the mass of the vehicle, $n_r \in \mathbb{N}$ denotes the number of rotors, $A \in \mathbb{R}$ corresponds to the rotor disk area, $f \in \mathbb{R}$ represents the equivalent flat plate area of the fuselage, $\kappa \in \mathbb{R}$ is the induced power correction factor, and $\varrho \in \mathbb{R}$ denotes the air density.

Hence, the optimization problem~\eqref{eq:optimizationProblemMotionPrimitives} can be reformulated by including an energy term to prioritize solutions where the forward speed along the trajectory is as close as possible to the optimal value $v^\star$ in terms of energy consumption. Namely, we write: \vspace{-1em}
\begin{equation}\label{eq:optimizationProblemEnergy}
    \resizebox{1\hsize}{!}{$%
    \begin{split}
    &\maximize_{\mathbf{p}^{(j)}, \, \mathbf{v}^{(j)}, \, \mathbf{a}^{(j)} \atop d \in \mathcal{D}} \;\;
    {\tilde{\rho}_\varphi (\mathbf{p}^{(j)}, \mathbf{v}^{(j)} ) -  \, \eta \sum_k \left( 1 - \frac{{^d}v_\mathrm{for}({^d}v_k)}{v^\star} \right)^2   } \\
    &\qquad \text{s.t.}~\quad\;\,\, {^d}\underline{v}^{(j)} \leq {^d}v^{(j)}_k \leq {^d}\bar{v}^{(j)}, \\
    &\,\;\;\;\,\, \qquad \quad\;\;\, {^d}\underline{a}^{(j)} \leq {^d}a^{(j)}_k \leq {^d}\bar{a}^{(j)}, \; \\
    %
    %
    &\,\;\;\;\;\, \qquad \quad\;\;\, \tilde{\rho}_\varphi (\mathbf{p}^{(j)}, \mathbf{v}^{(j)} ) \geq \varepsilon, \\
    &\,\;\;\;\;\, \qquad \quad\;\;\, {^d}\mathbf{S}^{(j)}, \forall k=\{0,1, \dots, N-1\},
    \end{split},
    $}%
\end{equation}
where $\eta \in [0,1] \in \mathbb{R}$ serves as a tunable weight for the energy term in the objective function, which quantifies the deviation of the forward speed ${^d}v_\mathrm{for}({^d}v_k)$ from the optimal value $v^\star$. The forward speed ${^d}v_\mathrm{for}({^d}v_k)$ is computed as the Euclidean norm of the velocity components of drone $d$ along the $x$ and $y$ axes:
\begin{equation}\label{eq:velocityInfty}
{^d}v_\mathrm{for}({^d}v_k) = \sqrt{ \left( {^d}v_k^{(1)} \right)^2 + \left( {^d}v_k^{(2)} \right)^2 } .
\end{equation}

Note that, as demonstrated in~\cite{Franco2026JINT, BauersfeldRAL2022}, the energy expended by multi-rotors during ascent, descent, and hovering maneuvers is constant since these operations are performed at a constant speed. Therefore, in the context of the optimization problem~\eqref{eq:optimizationProblemEnergy}, we can simplify the computation by excluding the ${^d}v^{(3)}$ component from the forward speed calculation in~\eqref{eq:velocityInfty}, reducing the computational burden.

It is important to consider the scenario where $v^\star$ exceeds the maximum feasible vehicle speed ${^d}\bar{v}^{(j)}$. In such cases, we can reformulate the optimization problem~\eqref{eq:optimizationProblemEnergy} by replacing $v^\star$ with $\lVert {^d}\bar{v}^{(1)} + {^d}\bar{v}^{(2)} \rVert$, which represents the feasible forward speed closest to $v^\star$ as defined in~\eqref{eq:velocityInftyStar}.

The introduction of the new energy term in the objective function represents a quadratic error that measures the deviation of the vehicle speed along the trajectory from the speed that minimizes energy consumption. It is worth noting that the solutions generated by the energy-aware planner still yield feasible trajectories that satisfy all mission requirements. However, these trajectories strike a balance between energy consumption and potential trade-offs in smooth robustness score $\tilde{\rho}_\varphi(\mathbf{x})$.    



\section{Experimental Results}
\label{sec:Validation}

To validate and assess the effectiveness of our proposed planning approach, we conducted a series of simulations and experiments. Initially, numerical simulations were carried out using MATLAB, where the actual vehicle dynamics and trajectory tracking controller were not explicitly modeled. This allowed us to evaluate the planning algorithm's performance and gain valuable insights into its behavior. Subsequently, to further verify the feasibility of the generated trajectories and leverage the benefits of software-in-the-loop simulations~\cite{Baca2020mrs, Silano2019SMC}, we conducted additional simulations using the Gazebo robotics simulator. Finally, field experiments performed in a mock-up scenario that closely resembles real-world conditions demonstrated the practical applicability of the proposed method. 

These experiments aimed to showcase several aspects: (i) the adherence of the planned trajectories to the mission requirements, (ii) the necessity of the~\ac{STL} motion planner in meeting the mission specifications, as the~\ac{MILP} formulation alone is insufficient, (iii) the capability of our approach to handle~\ac{UAV} failures and perform mission replanning, (iv) a comparison of the solutions obtained with and without the energy-aware feature, and (v) the feasibility of our method in real-world scenarios.

The optimization algorithm was implemented using MATLAB R2019b, with the~\ac{MILP} formulated using the CVX framework\footnote{\url{http://cvxr.com/cvx/}}, and the~\ac{STL} motion planner implemented using the CasADi library\footnote{\url{https://web.casadi.org/}} with IPOPT\footnote{\url{https://coin-or.github.io/Ipopt/}} as the solver. Within CVX, the choice of heuristic for solving the~\ac{MILP} problem was left to the framework itself. All numerical simulations and experiments were conducted on a computer equipped with an i7-8565U processor (\SI{1.80}{\giga\hertz}) and 32GB of RAM, running on the Ubuntu 20.04 operating system. For more detailed information and visual demonstrations of the experimental results, we provide videos that can be accessed at \url{http://mrs.felk.cvut.cz/bird-diverter-ar}.



\subsection{Bird diverters installation}
\label{sec:birdDiverterInstallationResults}

The proposed planning strategy was tested in a bird diverter installation scenario with two~\acp{UAV} operating in a mock-up scenario ($\SI{50}{\meter} \times \SI{20}{\meter} \times \SI{15}{\meter}$). The scenario had seven target regions and four refilling stations, as shown in Figure~\ref{fig:scenario}. The parameters used for the optimization problem are listed in Table~\ref{tab:tableParamters}. To avoid long waiting periods, symbolic values were used for the installation time $T_\mathrm{ins}$ and refilling time $T_\mathrm{rs}$. Heading angles were adjusted to align the~\acp{UAV} with the displacement direction or the cables during installation. It is assumed that an onboard low-level controller, e.g.~\cite{Armengol2021AIRPHARO, Suarez2021AIRPHARO, CacaceDrones2023}, handles the installation operation once the~\ac{UAV} reaches a region.

\begin{table}[tb]
	\centering
	\begin{adjustbox}{max width=1\columnwidth}
		\begin{tabular}{|c|c|c|c|c|c|}
		  \hline
		  \textbf{Parameter} & \textbf{Symbol} & \textbf{Value} & \textbf{Parameter} & \textbf{Symbol} & \textbf{Value}\\
		  \hline
		  Payload capacity (UAV1) & ${^1}\bar{c}$  & $\SI{2}{[-]}$ &
		  Payload capacity (UAV2) & ${^2}\bar{c}$ & $\SI{3}{[-]}$ \\
		  Payload capacity (UAV3) & ${^3}\bar{c}$ & $\SI{1}{[-]}$ &
            Payload capacity (UAV4) & ${^4}\bar{c}$ & $\SI{4}{[-]}$ \\
		  \ac{STL} safety margin & $\varepsilon$ & $\SI{0.2}{[-]}$ &
		  Safety mutual distance & $\Gamma_\mathrm{dis}$ & $\SI{3}{[\meter]}$ \\
		  Max. velocity & ${^d}\bar{v}^{(j)}$ & $\SI{3.1}{[\meter\per\second]}$ &
		  Max. acceleration & ${^d}\bar{a}^{(j)}$ & $\SI{3.1}{[\meter\per\square\second]}$ \\
		  Optimal forward speed & $v^\star$ & $\SI{2.5}{[\meter\per\second]}$ &
            Time replanning & $T_\mathrm{rep}$ & $\SI{10}{[\second]}$ \\
            Mission time & $T_N$ & $\SI{155}{[\second]}$ &
            Installation time & $T_\mathrm{ins}$ & $\SI{5}{[\second]}$ \\
            Refilling time & $T_\mathrm{rs}$ & $\SI{12}{[\second]}$ &
            Sampling period & $T_s$ & $\SI{0.05}{[\second]}$ \\
            Number of samples & $N$ & $\SI{1360}{[-]}$ &
            %
            %
            Scaling factor & $\lambda$ & $\SI{10}{[-]}$ \\
            Weight energy term & $\eta$ & $\SI{1}{[-]}$ &
            - & - & - \\
		\hline
		\end{tabular}
	\end{adjustbox}
	\caption{Parameter values for the optimization problem.}
	\label{tab:tableParamters}
\end{table}

\begin{figure}[tb]
    \centering
    \adjincludegraphics[width=\columnwidth]{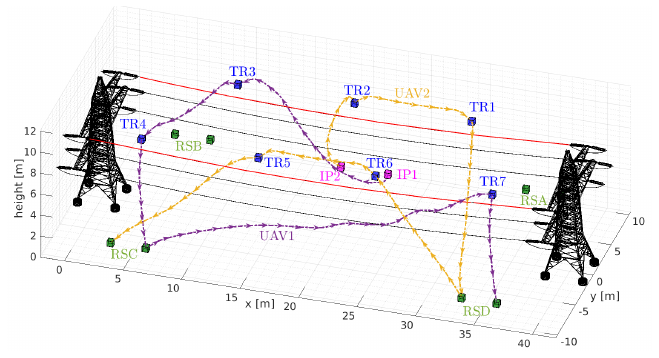}
    \caption{Bird diverter installation: planned trajectories for two~\acp{UAV}, seven target regions, and four refilling stations. The arrows indicate the paths followed by the~\acp{UAV} throughout the mission.}
    \label{fig:planner_base}
\end{figure}

\begin{figure}[tb]
    \hspace{-1.15em}
    \adjincludegraphics[width=\columnwidth 
    ]{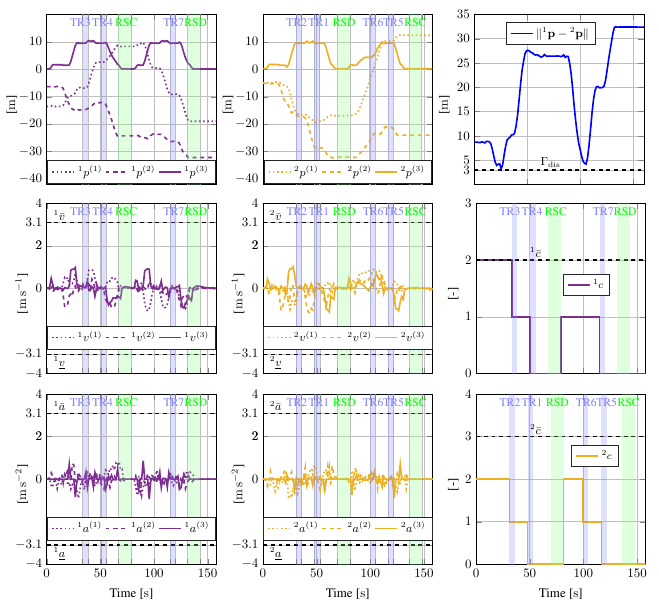}
    \caption{Position, velocity, acceleration, safety mutual distance, and payload capacity for ``\ac{UAV}1'' and ``\ac{UAV}2'' from real-word experiments. The installation and refilling time windows are highlighted in blue and green, respectively.}
    \label{fig:scenario_9RealWorldExp}
\end{figure}

\begin{figure}[tb]
    \centering
    \adjincludegraphics[width=\columnwidth 
    ]{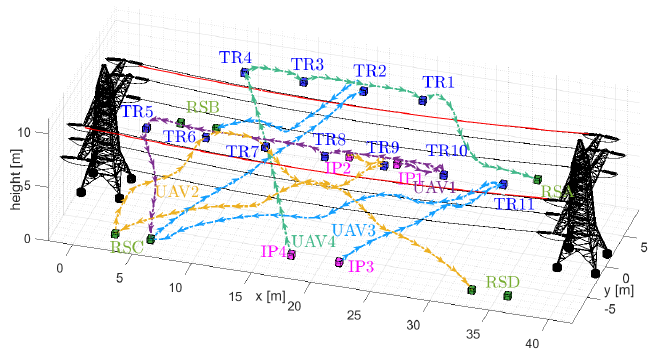}
    \caption{Bird diverter installation: planned trajectories for four~\acp{UAV}, eleven target regions, and four refilling stations. The arrows indicate the paths followed by the~\acp{UAV} throughout the mission.}
    \label{fig:planner_base_moreDrones}
\end{figure}

Figure~\ref{fig:planner_base} illustrates the planned trajectories, depicting the power towers, cables, target regions, refilling stations, and vehicles' initial position. The towers have a height of $\SI{15}{\meter}$ and are spaced $\SI{40}{\meter}$ apart. The optimization problem required $\SI{14}{\minute}$ to solve and $\SI{37}{\second}$ to find an initial guess solution. Real-world experiments demonstrate the compliance of the planned trajectories with mission requirements, as shown in Figure~\ref{fig:scenario_9RealWorldExp}. The figure confirms that the distance between vehicles always exceeds the threshold value $\Gamma_\mathrm{dis}$, the velocity and acceleration of the vehicles remain within the allowable bounds ($[{^d}\underline{v}^{(j)}, {^d}\bar{v}^{(j)}]$ and $[{^d}\underline{a}^{(j)}, {^d}\bar{a}^{(j)}]$), and the vehicles never visit a target region when they run out of diverters. Note that, for simplicity, the velocity and acceleration bounds are assumed to be symmetric, i.e., $\lvert {^d}\underline{v}^{(j)} \rvert = \lvert {^d}\bar{v}^{(j)} \rvert$ and $\lvert {^d}\underline{a}^{(j)} \rvert = \lvert {^d}\bar{a}^{(j)} \rvert$. Lastly, for comparative purposes, Figure~\ref{fig:planner_base_moreDrones} displays the planned trajectories for a more complex scenario involving four~\acp{UAV} and eleven target regions. In this case, the optimization problem required $\SI{22}{\minute}$ to solve and $\SI{57}{\second}$ to find an initial guess solution. This scenario offers an initial glimpse into the scalability and performance of the proposed approach with increased complexity and clutter.

\subsection{Comparison with the initial guess solution}
\label{sec:initialGuessResults}

\begin{figure}[tb]
    \centering
    \adjincludegraphics[width=\columnwidth 
    ]{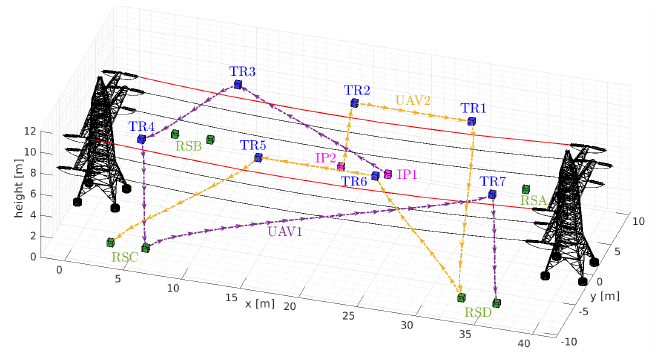}
    \caption{Bird diverter installation utilizing only the trajectories obtained from solving the~\ac{MILP} formulation, without employing the~\ac{STL} planner.}
    \label{fig:scenario_5}
\end{figure}

As outlined in Section~\ref{sec:problemFormulation}, the~\ac{MILP} formulation serves as an initial seed for the~\ac{STL} optimization problem, enabling the generation of feasible trajectories for the bird diverter installation mission. While the~\ac{MILP} solution is computationally efficient, it is insufficient on its own due to the omission of critical nonlinear aspects such as obstacle avoidance, safety distance requirements, and time constraints. It is important to note that the~\ac{STL} optimization problem, when not seeded with the~\ac{MILP} initial guess, does not converge to a feasible solution. Therefore, a direct comparison between the~\ac{STL}~\eqref{eq:optimizationProblemMotionPrimitives} and the~\ac{MILP}~\eqref{eq:MILP} solutions cannot be made. Nonetheless, we evaluate the compliance of both solutions with the mission requirements and assess how the proposed hierarchical planner addresses nonlinear complexities such as obstacle avoidance, safety distance, and time requirements, which the~\ac{MILP} alone cannot handle. Furthermore, we demonstrate how the subsequent~\ac{STL} optimization refines the initial solution obtained from the~\ac{MILP} formulation, striving for maximum robustness and satisfying all mission specifications. Figure~\ref{fig:scenario_5} illustrates the dynamically feasible trajectories obtained using multi-rotor motion primitives~\cite{Silano2021RAL}, along with the sequences of waypoints assigned to each~\ac{UAV} by the~\ac{MILP} solution.

Upon initial examination, it is evident that the~\ac{MILP} formulation does not explicitly consider time, resulting in trajectories that may exceed the specified time limit $T_N$ and become impractical for real-world applications. This limitation also impacts the trajectory quality, as the~\ac{MILP} approach does not control the vehicles' motion through accelerations ${^d}\mathbf{a}^{(j)}$, but rather sets waypoint sequences for the~\acp{UAV} to follow, leading to sharp corners and spikes that are challenging to execute in practice. Such abrupt changes in direction can strain the actuators and compromise trajectory tracking, potentially causing deviations from mission specifications and safety requirements, including the risk of collisions with power towers or cables. Additionally, this approach can result in high energy consumption, adversely affecting the mission objectives. Another crucial distinction lies in the treatment of the mutual safety distance constraint (${^d}\varphi_\mathrm{dis}$). While the~\ac{STL} optimization problem incorporates this constraint by regulating velocities and accelerations, and thus the vehicles' positions, without compromising the optimal region visitation sequence, the~\ac{MILP} formulation does not account for vehicle dynamics to mitigate computational complexity. Consequently, adjusting the value of $\Gamma_\mathrm{dis}$~\eqref{subeq:keepDistance} may require entirely different optimal sequences (outputs of the~\ac{MILP} solver) to obtain a feasible solution for the problem.

\begin{figure}[tb]
    \centering
    \adjincludegraphics[width=\columnwidth 
    ]{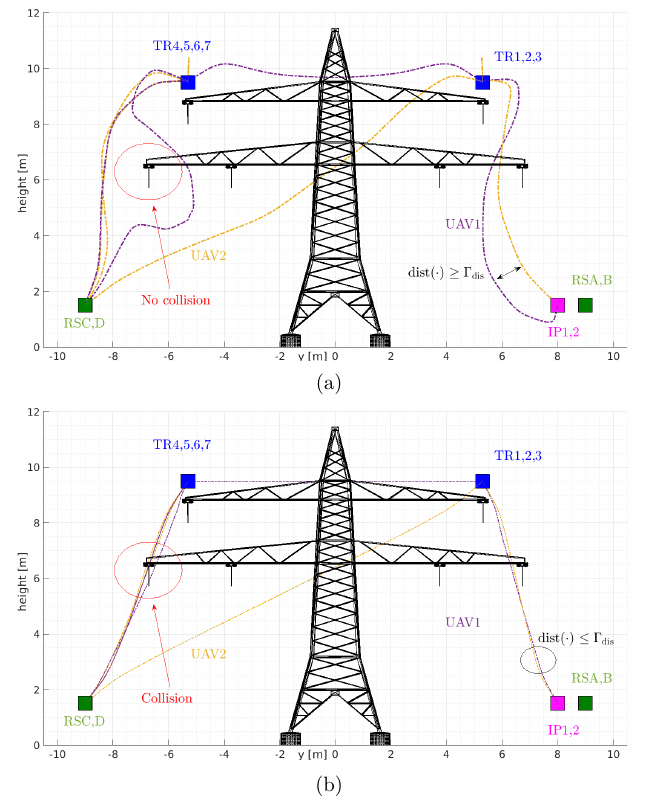}
    \caption{Comparison of trajectories obtained from the complete~\ac{STL} optimization 
    problem (\ref{fig:scenario_7}(a)) and the sole~\ac{MILP} formulation 
    (\ref{fig:scenario_7}(b)).}
    \label{fig:scenario_7}
\end{figure}

Figure~\ref{fig:scenario_7} provides a visual comparison between the solutions obtained by solely solving the~\ac{MILP} formulation and the complete~\ac{STL} optimization problem. The figure clearly illustrates how the~\ac{STL} formulation refines the initial guess solution from the~\ac{MILP}, resulting in trajectories that satisfy obstacle avoidance, safety distance requirements, and other mission specifications. Notably, the trajectories are significantly rearranged to achieve higher robustness values $\tilde{\rho}_\varphi(\mathbf{x})$.

\begin{figure}[tb]
    \centering
    \adjincludegraphics[width=\columnwidth]{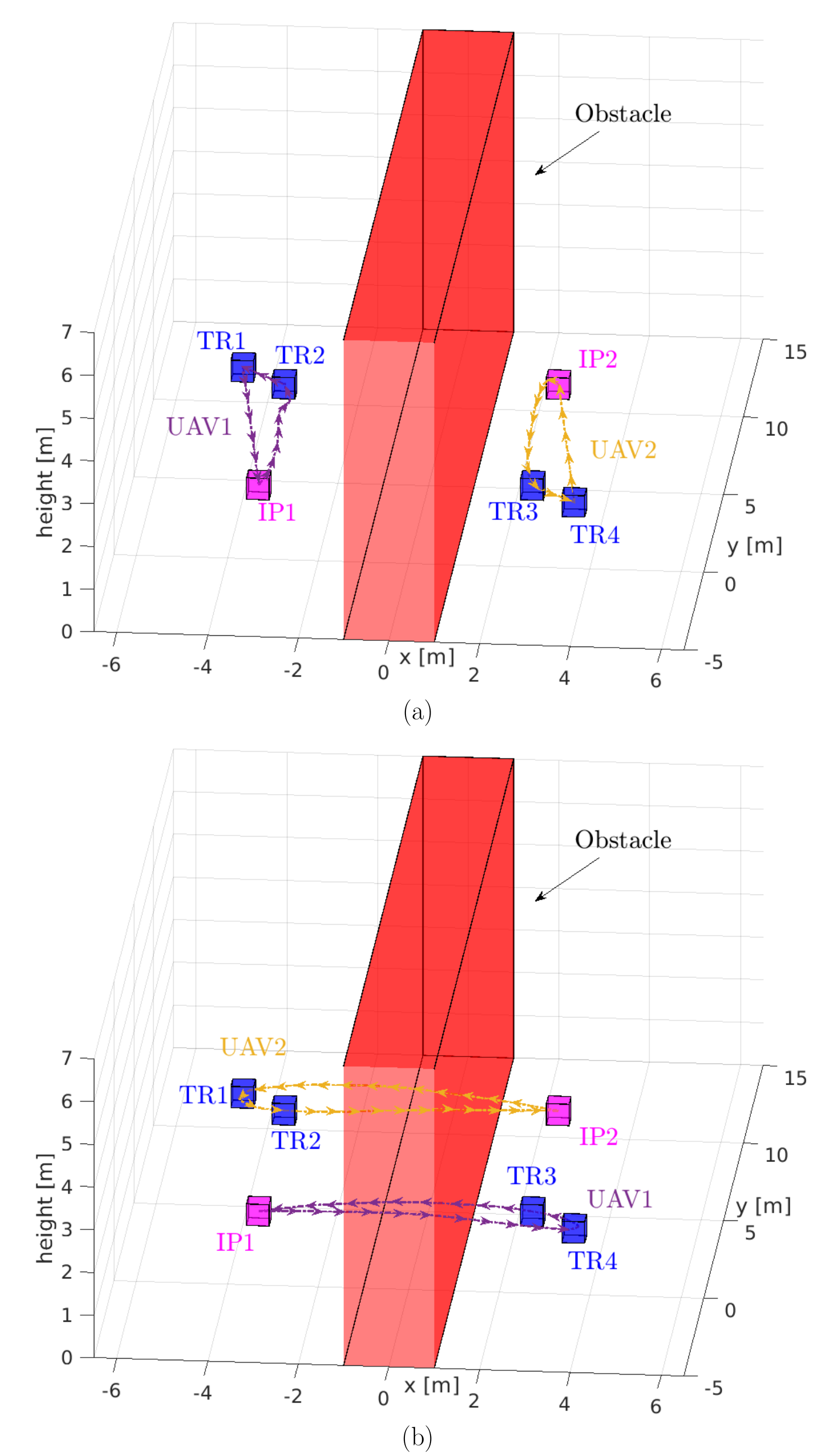}
    \caption{Simple scenario showing the independence of the final~\ac{STL} solution 
    (\ref{fig:toyExample}(a)) from the~\ac{MILP} initial guess (\ref{fig:toyExample}(b)).}
    \label{fig:toyExample}
\end{figure}

Moreover, it is important to highlight that the differences between the~\ac{STL} and~\ac{MILP} solutions extend beyond the trajectory shape and also encompass the sequence of target regions to visit. Figure~\ref{fig:toyExample} demonstrates this distinction in a simple scenario involving two~\acp{UAV} (located at ``IP1'' and ``IP2'') with identical dynamic constraints. The objective is for the~\acp{UAV} to visit a set of target regions (``TR1'', ``TR2'', ``TR3'', and ``TR4'') within the time interval $[0, T_N]$, while adhering to workspace constraints (${^d}\varphi_\mathrm{ws}$), avoiding obstacles (${^d}\varphi_\mathrm{obs}$), and maintaining a safe distance from each other (${^d}\varphi_\mathrm{dis}$). In Figure~\ref{fig:toyExample}(b), the initial guess solution from the~\ac{MILP} formulation assigns target regions based on workload balancing and minimizing travel distance, without considering obstacle avoidance. In contrast, the complete~\ac{STL} planner (Figure~\ref{fig:toyExample}(a)) reassigns the targets to satisfy all mission specifications. This example highlights the independence of the final~\ac{STL} optimization from the~\ac{MILP} initial guess solution, providing added flexibility to the hierarchical planner. However, it is essential to note that as the problem complexity increases, a significant deviation between the initial guess and the global solution may hinder the~\ac{STL} optimizer, resulting in convergence to local optima.

\subsection{Energy-aware and replanning strategy}
\label{sec:energyResults}

\begin{figure}[tb]
    \centering
    \adjincludegraphics[width=\columnwidth 
    ]{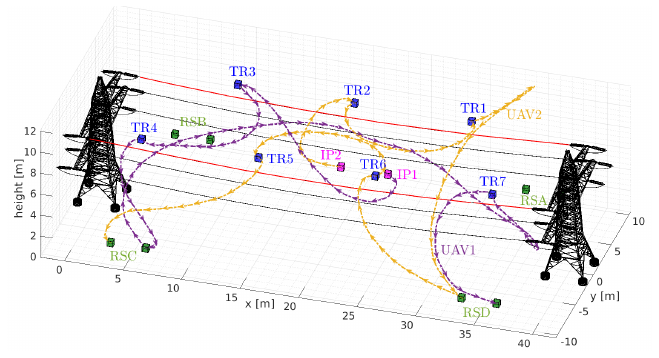}
    \caption{Bird diverter installation with trajectories obtained using the energy-aware planner.}
    \label{fig:scenario_4}
\end{figure}

\begin{figure}[tb]
    \centering
    \adjincludegraphics[width=\columnwidth 
    ]{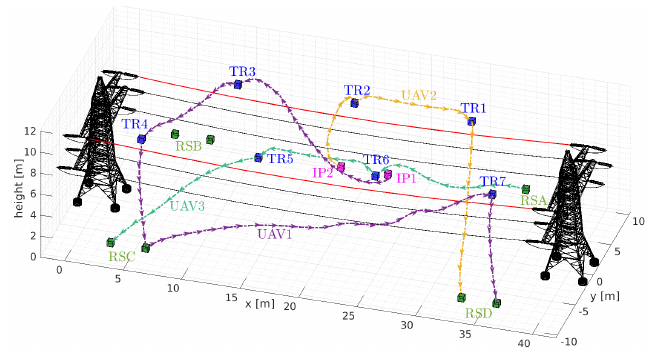}
    \caption{Bird diverter installation in the event of a ``UAV2'' failure, with the backup vehicle ``UAV3'' following the green path.}
    \label{fig:scenario_3}
\end{figure}

This section demonstrates the effectiveness of the energy-aware planner and event-triggered replanner, discussed in Sections~\ref{sec:energyAwarePlanner} and~\ref{sec:replanner}, respectively, in reducing energy consumption by the~\acp{UAV} and ensuring mission continuity in the event of~\ac{UAV} failures. The results of numerical simulations conducted in MATLAB are presented in Figures~\ref{fig:scenario_4} and~\ref{fig:scenario_3}.

\begin{figure}[tb]
    \centering
    \adjincludegraphics[width=\columnwidth 
    ]{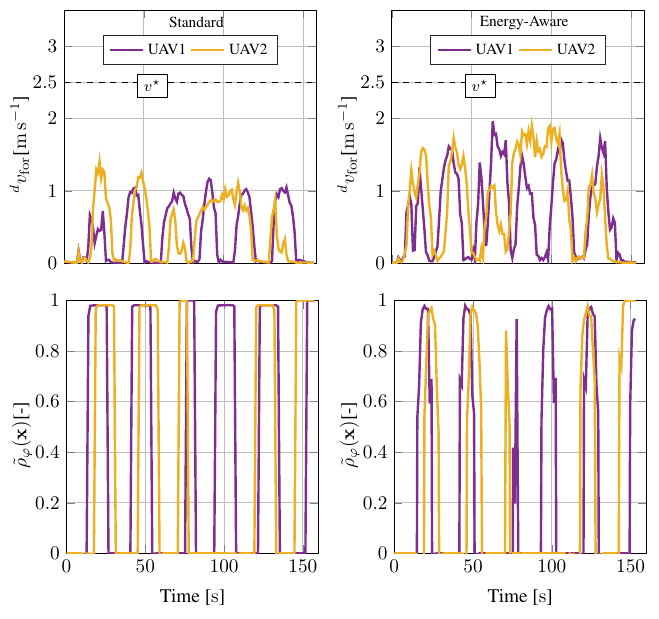}
    \caption{From left to right: comparison of forward speed ${^d}v_\mathrm{for}({^d}v_k)$ and normalized smooth robustness values $\tilde{\rho}_\varphi(\mathbf{x})$ for the standard and energy-aware planners.}
    \label{fig:scenario_6}
\end{figure}

Regarding the energy-aware planner, the trajectories incorporating the forward speed term $v^\star$ (Figure~\ref{fig:scenario_4}) exhibit longer paths and abrupt changes in direction compared to the trajectories without energy requirements (Figures~\ref{fig:planner_base} and~\ref{fig:planner_base_moreDrones}). This is due to the~\acp{UAV} needing to approach the energetically optimal forward speed $v^\star$ while satisfying the mission specifications $\varphi$ (eq.~\eqref{eq:longRangeInspection}) during the installation of diverters. As a result, the optimization problem~\eqref{eq:optimizationProblemEnergy} restricts the range of allowable velocities more than the standard planner~\eqref{eq:optimizationProblemMotionPrimitives}, where velocity values can be adjusted over a wider range to maximize robustness. Figure~\ref{fig:scenario_6} provides a comparison of the forward speed ${^d}v_\mathrm{for}$ and robustness values $\tilde{\rho}_\varphi(\mathbf{x})$ for the standard and energy-aware planners. The plot illustrates that higher velocities are associated with a slight decrease in smooth robustness values.

To demonstrate the event-triggered replanner, we conducted a scenario with a pair of~\acp{UAV} involved in the installation of bird diverters. Initially, both~\acp{UAV} followed the trajectories shown in Figure~\ref{fig:planner_base}. Subsequently, we simulated a failure event when ``UAV2'' (Figure~\ref{fig:scenario_3}) visited a refilling station to replenish diverters. This triggered the execution of the optimization problem~\eqref{eq:optimizationProblemMotionPrimitives} and the corresponding~\ac{MILP}~\eqref{eq:MILP} for the backup~\ac{UAV}, ``UAV3'', stationed at the refilling station ``RSA''. The optimization problem considered the remaining unvisited regions from~\ac{UAV}'s mission, ``UAV2'', and the trajectory of the non-faulty~\ac{UAV}, ``UAV1''. The resulting trajectories are depicted in Figure~\ref{fig:scenario_3}. In this particular scenario, the order of visiting the target regions remained the same, with slight variations observed in the trajectory connecting ``TR5'' and ``RSC'' (Figure~\ref{fig:planner_base}). These trajectory changes were driven by the optimization problem to ensure compliance with safety and obstacle avoidance requirements. The optimization process took approximately $\SI{2}{\second}$ to solve the entire problem, while the~\ac{MILP} solver took only a few tenths of a second. The failure event was detected at $T_\mathrm{fail}=\SI{9}{\second}$, considering a maximum expected computation time of $T_\mathrm{rep}=\SI{10}{\second}$.



\subsection{Field experiments}
\label{sec:fieldExperiments}

Experiments were conducted to evaluate the practical application of the proposed motion planning approach using DJI F450 quadrotors~\cite{MRS2022ICUAS_HW, MRS2023JINT_HW} in a mock-up scenario. The scenario, shown in Figure~\ref{fig:planner_base}, involved two~\acp{UAV} operating in a workspace that included power towers, cables, and refilling stations. To facilitate experimentation and consider the safety-critical nature of the power line environment, we simulated these objects in MATLAB and validated the trajectories in Gazebo. Videos showcasing simulated and experimental results can be accessed at \url{http://mrs.felk.cvut.cz/bird-diverter-ar}.

\begin{figure}[tb]
    \centering
    \adjincludegraphics[width=\columnwidth 
    ]{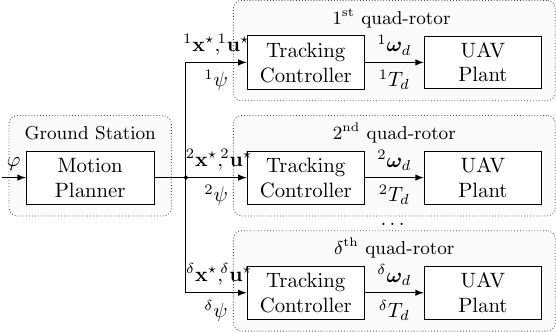}
    \caption{System Architecture. The \textit{STL Motion Planner} on the ground station generates trajectories $( {^1}\mathbf{x}^\star, {^1}\mathbf{u}^\star, \dots, {^\delta}\mathbf{x}^\star, {^\delta}\mathbf{u}^\star )$ and heading angles $( {^1}\psi, \dots, {^\delta}\psi )$ for the multi-rotors. These trajectories and heading angles are used as inputs to the \textit{Tracking Controller}, which computes thrust $( {^1}T_d, \dots, {^\delta}T_d )$ and angular velocities $( {^1}\bm{\omega}_d, \dots, {^\delta}\bm{\omega}_d )$ for the \textit{UAV Plant}~\cite{Silano2021RAL}}
    \label{fig:controlArchitecture}
\end{figure}

The system architecture, depicted in Figure~\ref{fig:controlArchitecture}, incorporates the~\ac{STL} motion planner, which solves the optimization problem~\eqref{eq:optimizationProblemMotionPrimitives} to generate trajectories (${^d}\mathbf{x}^\star$, ${^d}\mathbf{u}^\star$) and heading angles (${^d}\psi$) for the multi-rotors. The trajectory generation process is performed as a one-shot computation at time $t_0$, and the resulting trajectories serve as references for the~\ac{UAV} trajectory tracking controller~\cite{Baca2020mrs}. The~\acp{UAV} were equipped with an Intel NUC computer (i7-8559U processor with 16GB of RAM) and the Pixhawk flight controller. The software stack utilized the Noetic Ninjemys release of ROS running on Ubuntu 20.04. Further details can be found in~\cite{MRS2022ICUAS_HW, MRS2023JINT_HW}.

\begin{figure*}[tb]
    \centering
    \adjincludegraphics[width=2\columnwidth 
    ]{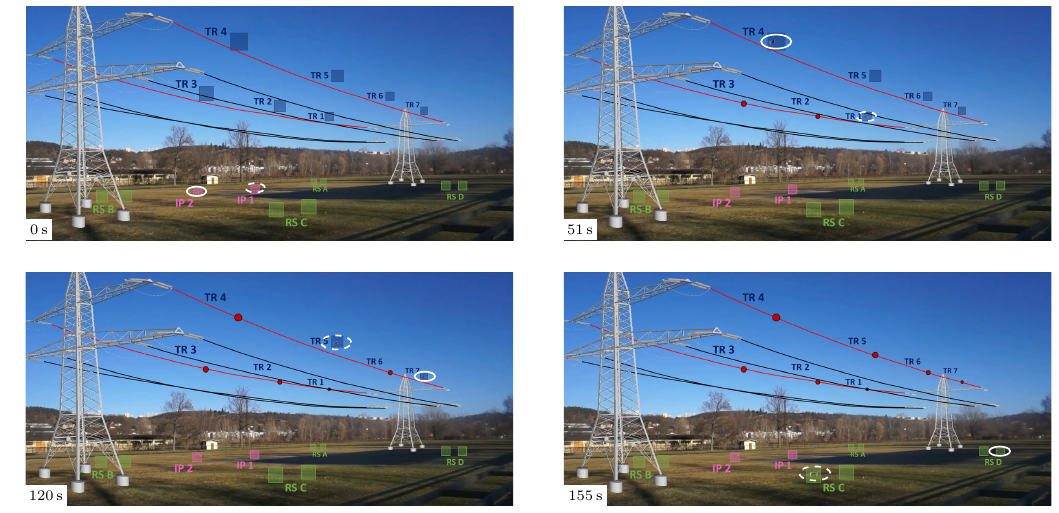}
    \caption{Field experiment snapshots showing the system evolution at different time 
    instants $t_k$. The positions of ``UAV1'' and ``UAV2'' are represented by solid and dashed 
    white lines, respectively. Power towers and cables are virtually projected onto the frames 
    for improved visualization. Target regions, refilling stations, and initial positions of 
    the~\acp{UAV} are indicated by blue, green, and magenta squares, respectively. The already 
    visited regions are highlighted by red spheres in each snapshot.}
    \label{fig:miniatureExperiments}
\end{figure*}

During the real flight experiments, we were able to verify the successful completion of the installation mission defined by the~\ac{STL} formula~\eqref{eq:longRangeInspection}. The flights also demonstrated compliance with physical constraints and safety requirements, including maximum velocity (${^d}\bar{v}$), maximum acceleration (${^d}\bar{a}$), and the minimum safety distance ($\Gamma_\mathrm{dis}$), as well as the payload capacity of the vehicles, ${^d}\bar{c}$ (Table~\ref{tab:tableParamters}). Figure~\ref{fig:miniatureExperiments} shows snapshots of the experiments, where the proximity of the~\acp{UAV} to the mechanical infrastructure (cables and towers) and refilling locations can be observed. To provide this visual representation, we used a ROS package to project 3D mesh files, which were originally used in MATLAB and Gazebo to represent the scenario, onto the camera frames of the~\acp{UAV}.

\section{Discussion}
\label{sec:discussion}

In this discussion section, we aim to provide a comprehensive analysis and insights into our 
proposed planning approach in the context of installing bird diverters. The work presented 
here forms a part of the AERIAL-CORE European project$^1$. While we acknowledge the existence 
of established approaches in the literature, such as the kinodynamic 
RRT$^\star$~\cite{WebbICRA2013} and the~\ac{CVRP} formulation~\cite{Dorling2017IEEESMCS},  
which were not originally designed for multi-robot systems and do not explicitly consider time 
requirements like~\ac{STL}, it is important to clarify that our objective is not to challenge 
or claim superiority over these approaches. Rather, our focus is on leveraging~\ac{TL} 
techniques and applying them to a real-world use-case scenario.

One of the distinct advantages of using~\acp{TL} is its ability to capture complex temporal constraints and requirements in a formal and expressive manner. By combining natural language commands, temporal and Boolean operators, and task and motion planning, we can generate trajectories that fulfill mission objectives while considering the dynamics and constraints of the multi-rotor system. While the robustness values can indeed be computed using functions of the variables, the use of~\acp{TL} provides a more concise and intuitive framework for specifying and addressing these mission specifications.

In our work, we make a deliberate choice to relax certain constraints in the~\ac{MILP} formulation, including obstacle avoidance, workspace limits, safety distance requirements, and mission time intervals (Section~\ref{sec:MILP}). This decision was made based on several considerations. Firstly, including these constraints would introduce complex nonlinearities and motion discontinuities, making the~\ac{MILP} formulation more challenging and computationally expensive and leading to a non-convex optimization problem. Secondly, considering time constraints would require incorporating motion primitives, further complicating the problem formulation. Adding excessive complexity in this aspect would not be beneficial, but rather counterproductive, as our goal is to obtain solutions as quickly as possible. The~\ac{MILP} formulation serves as a foundation for the subsequent~\ac{STL} optimization problem, acting as a stepping stone for the~\ac{STL} motion planner. Its primary purpose is to enable the avoidance of local optima and facilitate the refinement of the trajectory generation process.

To highlight the significance of our approach, we compare the solutions obtained from the~\ac{MILP} formulation and the complete~\ac{STL} optimization problem. The results, as showcased in Figure~\ref{fig:scenario_7}, demonstrate the refinement achieved by the~\ac{STL} formulation, particularly in addressing mission requirements such as obstacle avoidance and safety distance. This comparison serves to illustrate the added value and effectiveness of our proposed solution in the context of bird diverter installation. Furthermore, we highlight in Section~\ref{sec:initialGuessResults} and visually depict in Figure~\ref{fig:toyExample} a simple scenario example where we explicitly demonstrate that the differences between the~\ac{STL} and~\ac{MILP} solutions extend beyond the trajectory shape, encompassing the sequence of target regions to visit.

Regarding the replanning of other vehicles' routes (Section~\ref{sec:replanner}), our focus is on providing a practical solution that ensures continuity of operations in hazardous scenarios where swift action is crucial. We describe an online event-based replanning procedure where a new plan is computed for the backup multi-rotor in the event of a failure, while the trajectories of operational multi-rotors remain unchanged to ensure continuity. The replanning process aims to generate a new trajectory for the backup multi-rotor within a specified time interval, considering factors such as expected computation time. While the evaluation of alternative re-planning strategies for other vehicles' routes could be an interesting avenue for future research, it is important to note that our paper specifically addresses the installation of bird diverters and prioritizes the practicality and continuity of operations in this specific context. Although it may not provide the optimal solution for all vehicles' routes, it offers a fast and effective means to replace a faulty multi-rotor with a backup vehicle.

It is important to note that our work is specifically tailored to the practical context of installing bird diverters, which introduces unique challenges and constraints. These characteristics make our work novel and unique, as we have incorporated~\ac{TL} techniques into a problem that has not been extensively explored from this perspective. While existing procedures may address similar problems, our focus on applying~\ac{TL} techniques to this specific real-world scenario adds a valuable contribution to the field.

\subsection{Challenges of multi-robot bird diversion}

In this paper, we have explored the advantages and potential of using a multi-rotors team for bird diversion. However, it is important to address the drawbacks and considerations associated with this approach.

Coordination among multiple multi-rotors can be a complex task, requiring sophisticated control algorithms and communication protocols to ensure proper synchronization and cooperation. Effective communication is crucial for coordination and information sharing, but it can be susceptible to interference, signal loss, or limited bandwidth, which may disrupt the exchange of critical data. Furthermore, the presence of multi-rotors in the vicinity of birds may introduce disturbances that can affect their behavior. The noise and visual presence of the vehicles can cause stress or alter the natural patterns of bird movement. It is crucial to assess and mitigate these potential disturbances to minimize adverse effects on the target birds and their habitats.

Cost is another important consideration. The scalability and economic feasibility of employing a multi-rotors team for bird diversion, especially considering the large number of power towers, need to be carefully evaluated. Further research is required to optimize resource allocation and operational strategies, aiming to minimize costs while maintaining effectiveness. Moreover, the impact of wind and extreme weather on drone trajectories should be taken into account. Strong winds can affect the stability and maneuverability of multi-rotor systems, potentially hindering their effectiveness in bird diversion tasks. Integrating weather forecasting and real-time adaptation mechanisms can enhance safety and performance under varying environmental conditions.




\section{Conclusions}
\label{sec:conclusions}

This paper has presented a motion planning framework for encoding bird diverter installation missions for a team of multi-rotors with payload capacity limitations and dynamic constraints. The proposed method utilizes~\ac{STL} specifications to generate dynamically feasible trajectories that satisfy mission requirements, including safety and mission time constraints. The work builds upon a previous motion planner by adopting an~\ac{MILP} approach to handle the nonlinear non-convex optimization problem. The~\ac{MILP} solver provides an initial guess solution for the~\ac{STL} framework, facilitating algorithm convergence. Additionally, event-triggered replanning and energy-aware planning techniques are introduced to address~\ac{UAV} failures and optimize energy consumption, respectively. MATLAB and Gazebo simulations, along with field experiments, validate the effectiveness of the proposed approach in a real-world scenario.

Our analysis reveals that the simplified~\ac{MILP} solver alone is insufficient for the application, but it serves as a valuable initial guess for the complete~\ac{STL} planner. The hierarchical approach allows us to address a broader range of mission specifications and requirements compared to existing methods, albeit with increased computational complexity. Future work will focus on reducing this computational burden and enhancing scalability by incorporating decentralized feedback control laws based on time-varying control barrier functions. Additionally, exploring conflicting temporal logic specifications and other temporal logic languages will enable the application of the framework in dynamically changing environments.



\balance

\backmatter


\balance

\backmatter

\bmhead{Acknowledgments}

The authors would like to thank Daniel Smrcka, Jan Bednar, Jiri Horyna, Tomas Baca, Martin Saska and the MRS group in Prague for their help with the field experiments. The authors also would like to thank Jesus Capitan for his advice and valuable feedback to the development of the technical part.



\subsection*{Declarations}

\bmhead{Conflict of interest}

The authors declare that they have no known competing financial interests or personal relationships that could have appeared to influence the work reported in this paper.


\bmhead{Funding}

This publication is part of the R+D+i project TED2021-131716B-C22, funded by MCIN/AEI/10.13039/501100011033 and by the European Union NextGenerationEU/PRTR. This work was also supported by the European Union’s Horizon 2020 research and innovation project AERIAL-CORE under grant agreement no. 871479.



\bmhead{Author Contributions}

\textbf{Alvaro Caballero}: Conceptualization, Methodology, Software, Formal analysis, Writing – original draft. \textbf{Giuseppe Silano}: Conceptualization, Methodology, Software, Formal analysis, Writing – original draft, review \& editing, Supervision.



\bibliographystyle{bst/sn-mathphys.bst}
\bibliography{bib.bib}


\begin{thebibliography}{67}
\ifx \bisbn   \undefined \def \bisbn  #1{ISBN #1}\fi
\ifx \binits  \undefined \def \binits#1{#1}\fi
\ifx \bauthor  \undefined \def \bauthor#1{#1}\fi
\ifx \batitle  \undefined \def \batitle#1{#1}\fi
\ifx \bjtitle  \undefined \def \bjtitle#1{#1}\fi
\ifx \bvolume  \undefined \def \bvolume#1{\textbf{#1}}\fi
\ifx \byear  \undefined \def \byear#1{#1}\fi
\ifx \bissue  \undefined \def \bissue#1{#1}\fi
\ifx \bfpage  \undefined \def \bfpage#1{#1}\fi
\ifx \blpage  \undefined \def \blpage #1{#1}\fi
\ifx \burl  \undefined \def \burl#1{\textsf{#1}}\fi
\ifx \doiurl  \undefined \def \doiurl#1{\url{https://doi.org/#1}}\fi
\ifx \betal  \undefined \def \betal{\textit{et al.}}\fi
\ifx \binstitute  \undefined \def \binstitute#1{#1}\fi
\ifx \binstitutionaled  \undefined \def \binstitutionaled#1{#1}\fi
\ifx \bctitle  \undefined \def \bctitle#1{#1}\fi
\ifx \beditor  \undefined \def \beditor#1{#1}\fi
\ifx \bpublisher  \undefined \def \bpublisher#1{#1}\fi
\ifx \bbtitle  \undefined \def \bbtitle#1{#1}\fi
\ifx \bedition  \undefined \def \bedition#1{#1}\fi
\ifx \bseriesno  \undefined \def \bseriesno#1{#1}\fi
\ifx \blocation  \undefined \def \blocation#1{#1}\fi
\ifx \bsertitle  \undefined \def \bsertitle#1{#1}\fi
\ifx \bsnm \undefined \def \bsnm#1{#1}\fi
\ifx \bsuffix \undefined \def \bsuffix#1{#1}\fi
\ifx \bparticle \undefined \def \bparticle#1{#1}\fi
\ifx \barticle \undefined \def \barticle#1{#1}\fi
\bibcommenthead
\ifx \bconfdate \undefined \def \bconfdate #1{#1}\fi
\ifx \botherref \undefined \def \botherref #1{#1}\fi
\ifx \url \undefined \def \url#1{\textsf{#1}}\fi
\ifx \bchapter \undefined \def \bchapter#1{#1}\fi
\ifx \bbook \undefined \def \bbook#1{#1}\fi
\ifx \bcomment \undefined \def \bcomment#1{#1}\fi
\ifx \oauthor \undefined \def \oauthor#1{#1}\fi
\ifx \citeauthoryear \undefined \def \citeauthoryear#1{#1}\fi
\ifx \endbibitem  \undefined \def \endbibitem {}\fi
\ifx \bconflocation  \undefined \def \bconflocation#1{#1}\fi
\ifx \arxivurl  \undefined \def \arxivurl#1{\textsf{#1}}\fi
\csname PreBibitemsHook\endcsname

\bibitem{EPRI2008TechnicalReport}
\begin{botherref}
\oauthor{\bsnm{Major}, \binits{J.}},
\oauthor{\bsnm{Alvarez}, \binits{J.}},
\oauthor{\bsnm{Franke}, \binits{E.}},
\oauthor{\bsnm{Light}, \binits{G.}},
\oauthor{\bsnm{Allen}, \binits{P.}},
\oauthor{\bsnm{Edwards}, \binits{S.}}:
{Future inspection of overhead transmission lines}.
Technical report,
Electric Power Research Institute
(2008).
No. 1016921
\end{botherref}
\endbibitem

\bibitem{Hunting2002CECOM}
\begin{bbook}
\bauthor{\bsnm{Hunting}, \binits{K.}}:
\bbtitle{{A Roadmap for PIER Research on Avian Collisions with Power Lines in
  California. Commission Staff Report}}.
\bpublisher{California Energy Commission}, \blocation{???}
(\byear{2002}).
\burl{https://tinyurl.com/3v6yrehm}
\end{bbook}
\endbibitem

\bibitem{Ferrer2020GEC}
\begin{barticle}
\bauthor{\bsnm{{Ferrer}}, \binits{M.}},
\bauthor{\bsnm{{Morandini}}, \binits{V.}},
\bauthor{\bsnm{{Baumbusch}}, \binits{R.}},
\bauthor{\bsnm{{Muriel}}, \binits{R.}},
\bauthor{\bsnm{{De Lucas}}, \binits{M.}},
\bauthor{\bsnm{{Calabuig}}, \binits{C.}}:
\batitle{{Efficacy of different types of “bird flight diverter” in reducing
  bird mortality due to collision with transmission power lines}}.
\bjtitle{Global Ecology and Conservation}
\bvolume{23},
\bfpage{01130}
(\byear{2020}).
\doiurl{10.1016/j.gecco.2020.e01130}
\end{barticle}
\endbibitem

\bibitem{Manville1999}
\begin{botherref}
\oauthor{\bsnm{Manville}, \binits{A.M.}}:
{The ABCs of avoiding bird collisions at communication towers: The next steps}.
In Proceedings of theWorkshop on Avian Interactions with Utility and
  Communication Structures.
Electric Power Research Institute: Palo Alto, CA, USA
(1999).
\url{https://shorturl.at/CFHJ6}
\end{botherref}
\endbibitem

\bibitem{Manville2008}
\begin{botherref}
\oauthor{\bsnm{Manville}, \binits{A.M.}}:
{Towers, turbines, power lines, and buildings–steps being taken by the US
  Fish and Wildlife Service to avoid or minimize take of migratory birds at
  these structures}.
In Proceedings of theWorkshop on Avian Interactions with Utility and
  Communication Structures.
In Proceedings of the Fourth International Partners in Flight Conference:
  Tundra to Tropics, McAllen, TX, USA
(2008).
\url{https://shorturl.at/pFX39}
\end{botherref}
\endbibitem

\bibitem{Jenkins2010BCI}
\begin{barticle}
\bauthor{\bsnm{Jenkins}, \binits{A.R.}},
\bauthor{\bsnm{Smallie}, \binits{J.J.}},
\bauthor{\bsnm{Diamond}, \binits{M.A.}}:
\batitle{{Collisions with power lines: A global review of causes and mitigation
  with a South African perspectives}}.
\bjtitle{Bird Conservation International}
\bvolume{20},
\bfpage{263}--\blpage{278}
(\byear{2010})
\end{barticle}
\endbibitem

\bibitem{Ferrer2020}
\begin{botherref}
\oauthor{\bsnm{Ferrer}, \binits{M.}}:
{Birds and Power Lines: From Conflict to Solution}.
ENDESA SA and Fundacion MIGRES
(2020).
\url{https://shorturl.at/pAIJY}
\end{botherref}
\endbibitem

\bibitem{Suarez2021AIRPHARO}
\begin{bchapter}
\bauthor{\bsnm{Suarez}, \binits{A.}},
\bauthor{\bsnm{Romero}, \binits{H.}},
\bauthor{\bsnm{Salmoral}, \binits{R.}},
\bauthor{\bsnm{Acosta}, \binits{J.A.}},
\bauthor{\bsnm{Zambrano}, \binits{J.}},
\bauthor{\bsnm{Ollero}, \binits{A.}}:
\bctitle{{Experimental Evaluation of Aerial Manipulation Robot for the
  Installation of Clip Type Bird Diverters: Outdoor Flight Tests}}.
In: \bbtitle{Aerial Robotic Systems Physically Interacting with the
  Environment},
pp. \bfpage{1}--\blpage{7}
(\byear{2021}).
\doiurl{10.1109/AIRPHARO52252.2021.9571029}
\end{bchapter}
\endbibitem

\bibitem{Castano2021AppSci}
\begin{botherref}
\oauthor{\bsnm{Rodriguez-Castano}, \binits{A.}},
\oauthor{\bsnm{Nekoo}, \binits{S.R.}},
\oauthor{\bsnm{Romero}, \binits{H.}},
\oauthor{\bsnm{Salmoral}, \binits{R.}},
\oauthor{\bsnm{Acosta}, \binits{J.A.}},
\oauthor{\bsnm{Ollero}, \binits{A.}}:
{Installation of Clip-Type Bird Flight Diverters on High-Voltage Power Lines
  with Aerial Manipulation Robot: Prototype and Testbed Experimentation}.
Applied Sciences
\textbf{11}(16)
(2021).
\doiurl{10.3390/app11167427}
\end{botherref}
\endbibitem

\bibitem{CacaceDrones2023}
\begin{botherref}
\oauthor{\bsnm{Cacace}, \binits{J.}},
\oauthor{\bsnm{Giampetraglia}, \binits{L.}},
\oauthor{\bsnm{Ruggiero}, \binits{F.}},
\oauthor{\bsnm{Lippiello}, \binits{V.}}:
A novel gripper prototype for helical bird diverter manipulation.
Drones
\textbf{7}(1)
(2023).
\doiurl{10.3390/drones7010060}
\end{botherref}
\endbibitem

\bibitem{Suarez2021Access}
\begin{barticle}
\bauthor{\bsnm{Suarez}, \binits{A.}},
\bauthor{\bsnm{Salmoral}, \binits{R.}},
\bauthor{\bsnm{Zarco-Periñan}, \binits{P.J.}},
\bauthor{\bsnm{Ollero}, \binits{A.}}:
\batitle{{Experimental Evaluation of Aerial Manipulation Robot in Contact With
  15 kV Power Line: Shielded and Long Reach Configurations}}.
\bjtitle{IEEE Access}
\bvolume{9},
\bfpage{94573}--\blpage{94585}
(\byear{2021}).
\doiurl{10.1109/ACCESS.2021.3093856}
\end{barticle}
\endbibitem

\bibitem{Armengol2021AIRPHARO}
\begin{bchapter}
\bauthor{\bsnm{Armengol}, \binits{I.}},
\bauthor{\bsnm{Suarez}, \binits{A.}},
\bauthor{\bsnm{Heredia}, \binits{G.}},
\bauthor{\bsnm{Ollero}, \binits{A.}}:
\bctitle{{Design, Integration and Testing of Compliant Gripper for the
  Installation of Helical Bird Diverters on Power Lines}}.
In: \bbtitle{Aerial Robotic Systems Physically Interacting with the
  Environment},
pp. \bfpage{1}--\blpage{8}
(\byear{2021}).
\doiurl{10.1109/AIRPHARO52252.2021.9571044}
\end{bchapter}
\endbibitem

\bibitem{Afifi2022ICRA}
\begin{bchapter}
\bauthor{\bsnm{Afifi}, \binits{A.}},
\bauthor{\bparticle{van} \bsnm{Holland}, \binits{M.}},
\bauthor{\bsnm{Franchi}, \binits{A.}}:
\bctitle{{Toward Physical Human-Robot Interaction Control with Aerial
  Manipulators: Compliance, Redundancy Resolution, and Input Limits}}.
In: \bbtitle{International Conference on Robotics and Automation},
pp. \bfpage{4855}--\blpage{4861}
(\byear{2022}).
\doiurl{10.1109/ICRA46639.2022.9812451}
\end{bchapter}
\endbibitem

\bibitem{Afifi2023ICUAS}
\begin{bchapter}
\bauthor{\bsnm{Afifi}, \binits{A.}},
\bauthor{\bsnm{Corsini}, \binits{G.}},
\bauthor{\bsnm{Sable}, \binits{Q.}},
\bauthor{\bsnm{Aboudorra}, \binits{Y.}},
\bauthor{\bsnm{Sidobre}, \binits{D.}},
\bauthor{\bsnm{Franchi}, \binits{A.}}:
\bctitle{{Physical Human-Aerial Robot Interaction and Collaboration:
  Exploratory Results and Lessons Learned}}.
In: \bbtitle{International Conference on Unmanned Aircraft Systems},
pp. \bfpage{956}--\blpage{962}
(\byear{2023}).
\doiurl{10.1109/ICUAS57906.2023.10156609}
\end{bchapter}
\endbibitem

\bibitem{donze2010ICFMATS}
\begin{bchapter}
\bauthor{\bsnm{Donze}, \binits{A.}},
\bauthor{\bsnm{Maler}, \binits{O.}}:
\bctitle{{Robust satisfaction of temporal logic over real-valued signals}}.
In: \bbtitle{{International Conference on Formal Modeling and Analysis of Timed
  Systems}},
pp. \bfpage{92}--\blpage{106}
(\byear{2010}).
\doiurl{10.1007/978-3-642-15297-9\_9}.
\bcomment{Springer}
\end{bchapter}
\endbibitem

\bibitem{Pola2019ARC}
\begin{barticle}
\bauthor{\bsnm{{Pola}}, \binits{G.}},
\bauthor{\bsnm{{Di Benedetto}}, \binits{M.D.}}:
\batitle{{Control of Cyber-Physical-Systems with logic specifications: A formal
  methods approach}}.
\bjtitle{Annual Reviews in Control}
\bvolume{47},
\bfpage{178}--\blpage{192}
(\byear{2019}).
\doiurl{10.1016/j.arcontrol.2019.03.010}
\end{barticle}
\endbibitem

\bibitem{maler2004FTMATFTS}
\begin{bchapter}
\bauthor{\bsnm{Maler}, \binits{O.}},
\bauthor{\bsnm{Nickovic}, \binits{D.}}:
\bctitle{{Monitoring temporal properties of continuous signals}}.
In: \bbtitle{Formal Techniques, Modelling and Analysis of Timed and
  Fault-Tolerant Systems},
pp. \bfpage{152}--\blpage{166}.
\bpublisher{Springer},
\blocation{Berlin, Heidelberg}
(\byear{2004}).
\doiurl{10.1007/978-3-540-30206-3\_12}
\end{bchapter}
\endbibitem

\bibitem{karaman2011IJRR}
\begin{barticle}
\bauthor{\bsnm{Karaman}, \binits{S.}},
\bauthor{\bsnm{Frazzoli}, \binits{E.}}:
\batitle{{Sampling-based algorithms for optimal motion planning}}.
\bjtitle{The International Journal of Robotics Research}
\bvolume{30}(\bissue{7}),
\bfpage{846}--\blpage{894}
(\byear{2011}).
\doiurl{10.1177/0278364911406761}
\end{barticle}
\endbibitem

\bibitem{hauser_tro16}
\begin{barticle}
\bauthor{\bsnm{Hauser}, \binits{K.}},
\bauthor{\bsnm{Zhou}, \binits{Y.}}:
\batitle{{Asymptotically Optimal Planning by Feasible Kinodynamic Planning in a
  State-Cost Space}}.
\bjtitle{IEEE Transactions on Robotics}
\bvolume{32}(\bissue{6}),
\bfpage{1431}--\blpage{1443}
(\byear{2016}).
\doiurl{10.1109/TRO.2016.2602363}
\end{barticle}
\endbibitem

\bibitem{Webb2013ICRA}
\begin{bchapter}
\bauthor{\bsnm{Webb}, \binits{D.J.}},
\bauthor{\bparticle{van~den} \bsnm{Berg}, \binits{J.}}:
\bctitle{{Kinodynamic RRT*: Asymptotically optimal motion planning for robots
  with linear dynamics}}.
In: \bbtitle{IEEE International Conference on Robotics and Automation},
pp. \bfpage{5054}--\blpage{5061}
(\byear{2013}).
\doiurl{10.1109/ICRA.2013.6631299}
\end{bchapter}
\endbibitem

\bibitem{Ryll2019ICRA}
\begin{bchapter}
\bauthor{\bsnm{Ryll}, \binits{M.}},
\bauthor{\bsnm{Ware}, \binits{J.}},
\bauthor{\bsnm{Carter}, \binits{J.}},
\bauthor{\bsnm{Roy}, \binits{N.}}:
\bctitle{{Efficient Trajectory Planning for High Speed Flight in Unknown
  Environments}}.
In: \bbtitle{International Conference on Robotics and Automation},
pp. \bfpage{732}--\blpage{738}
(\byear{2019}).
\doiurl{10.1109/ICRA.2019.8793930}
\end{bchapter}
\endbibitem

\bibitem{chen2021RAL}
\begin{barticle}
\bauthor{\bsnm{Chen}, \binits{G.}},
\bauthor{\bsnm{Sun}, \binits{D.}},
\bauthor{\bsnm{Dong}, \binits{W.}},
\bauthor{\bsnm{Sheng}, \binits{X.}},
\bauthor{\bsnm{Zhu}, \binits{X.}},
\bauthor{\bsnm{Ding}, \binits{H.}}:
\batitle{{Computationally Efficient Trajectory Planning for High Speed Obstacle
  Avoidance of a Quadrotor With Active Sensing}}.
\bjtitle{IEEE Robotics and Automation Letters}
\bvolume{6}(\bissue{2}),
\bfpage{3365}--\blpage{3372}
(\byear{2021}).
\doiurl{10.1109/LRA.2021.3062332}
\end{barticle}
\endbibitem

\bibitem{Tan2021IEEEAccess}
\begin{barticle}
\bauthor{\bsnm{Tan}, \binits{C.S.}},
\bauthor{\bsnm{Mohd-Mokhtar}, \binits{R.}},
\bauthor{\bsnm{Arshad}, \binits{M.R.}}:
\batitle{{A Comprehensive Review of Coverage Path Planning in Robotics Using
  Classical and Heuristic Algorithms}}.
\bjtitle{IEEE Access}
\bvolume{9},
\bfpage{119310}--\blpage{119342}
(\byear{2021}).
\doiurl{10.1109/ACCESS.2021.3108177}
\end{barticle}
\endbibitem

\bibitem{Wang2023AI}
\begin{barticle}
\bauthor{\bsnm{Wang}, \binits{X.}},
\bauthor{\bsnm{Li}, \binits{B.}},
\bauthor{\bsnm{Su}, \binits{X.}},
\bauthor{\bsnm{Peng}, \binits{H.P.}},
\bauthor{\bsnm{Wang}, \binits{L.}},
\bauthor{\bsnm{Lu}, \binits{C.}},
\bauthor{\bsnm{Wang}, \binits{C.}}:
\batitle{{Autonomous dispatch trajectory planning on flight deck: A
  search-resampling-optimization framework}}.
\bjtitle{Engineering Applications of Artificial Intelligence}
\bvolume{119}(\bissue{105792}),
\bfpage{1}--\blpage{14}
(\byear{2023}).
\doiurl{10.1016/j.engappai.2022.105792}
\end{barticle}
\endbibitem

\bibitem{Wang2023OceanEngineering}
\begin{barticle}
\bauthor{\bsnm{Wang}, \binits{X.}},
\bauthor{\bsnm{Deng}, \binits{Z.}},
\bauthor{\bsnm{Peng}, \binits{H.}},
\bauthor{\bsnm{Wang}, \binits{L.}},
\bauthor{\bsnm{Wang}, \binits{Y.}},
\bauthor{\bsnm{Tao}, \binits{L.}},
\bauthor{\bsnm{Lu}, \binits{C.}},
\bauthor{\bsnm{Peng}, \binits{Z.}}:
\batitle{{Autonomous docking trajectory optimization for unmanned surface
  vehicle: A hierarchical method}}.
\bjtitle{Ocean Engineering}
\bvolume{279}(\bissue{114156}),
\bfpage{1}--\blpage{17}
(\byear{2023}).
\doiurl{10.1016/j.oceaneng.2023.114156}
\end{barticle}
\endbibitem

\bibitem{Gao2023AI}
\begin{barticle}
\bauthor{\bsnm{Gao}, \binits{X.}},
\bauthor{\bsnm{Wang}, \binits{L.}},
\bauthor{\bsnm{Yu}, \binits{X.}},
\bauthor{\bsnm{Su}, \binits{X.}},
\bauthor{\bsnm{Ding}, \binits{Y.}},
\bauthor{\bsnm{Lu}, \binits{H.} \bsuffix{Chen~Peng}},
\bauthor{\bsnm{Wang}, \binits{X.}}:
\batitle{{Conditional probability based multi-objective cooperative task
  assignment for heterogeneous UAVs}}.
\bjtitle{Engineering Applications of Artificial Intelligence}
\bvolume{123}(\bissue{106404}),
\bfpage{1}--\blpage{22}
(\byear{2023}).
\doiurl{10.1016/j.engappai.2023.106404}
\end{barticle}
\endbibitem

\bibitem{Nekovar2021RAL}
\begin{barticle}
\bauthor{\bsnm{{Nekovář}}, \binits{F.}},
\bauthor{\bsnm{{Faigl}}, \binits{J.}},
\bauthor{\bsnm{{Saska}}, \binits{M.}}:
\batitle{{Multi-Tour Set Traveling Salesman Problem in Planning Power
  Transmission Line Inspection}}.
\bjtitle{IEEE Robotics and Automation Letters}
\bvolume{6}(\bissue{4}),
\bfpage{6196}--\blpage{6203}
(\byear{2021}).
\doiurl{10.1109/LRA.2021.3091695}
\end{barticle}
\endbibitem

\bibitem{Xiong2022CCDC}
\begin{bchapter}
\bauthor{\bsnm{Xiong}, \binits{H.}},
\bauthor{\bsnm{Lei}, \binits{D.}},
\bauthor{\bsnm{Li}, \binits{M.}}:
\bctitle{{Multi-objective traveling salesman problem with drone: imperialist
  competitive algorithm}}.
In: \bbtitle{2022 34th Chinese Control and Decision Conference},
pp. \bfpage{3635}--\blpage{3640}
(\byear{2022}).
\doiurl{10.1109/CCDC55256.2022.10034189}
\end{bchapter}
\endbibitem

\bibitem{coutinho2018unmanned}
\begin{barticle}
\bauthor{\bsnm{Coutinho}, \binits{W.P.}},
\bauthor{\bsnm{Battarra}, \binits{M.}},
\bauthor{\bsnm{Fliege}, \binits{J.}}:
\batitle{{The unmanned aerial vehicle routing and trajectory optimisation
  problem, a taxonomic review}}.
\bjtitle{Computers \& Industrial Engineering}
\bvolume{120},
\bfpage{116}--\blpage{128}
(\byear{2018}).
\doiurl{10.1016/j.cie.2018.04.037}
\end{barticle}
\endbibitem

\bibitem{FaiglRAL2019}
\begin{barticle}
\bauthor{\bsnm{Faigl}, \binits{J.}},
\bauthor{\bsnm{Váňa}, \binits{P.}},
\bauthor{\bsnm{Deckerová}, \binits{J.}}:
\batitle{{Fast Heuristics for the 3-D Multi-Goal Path Planning Based on the
  Generalized Traveling Salesman Problem With Neighborhoods}}.
\bjtitle{IEEE Robotics and Automation Letters}
\bvolume{4}(\bissue{3}),
\bfpage{2439}--\blpage{2446}
(\byear{2019}).
\doiurl{10.1109/LRA.2019.2900507}
\end{barticle}
\endbibitem

\bibitem{penicka2019RAL}
\begin{barticle}
\bauthor{\bsnm{Penicka}, \binits{R.}},
\bauthor{\bsnm{Faigl}, \binits{J.}},
\bauthor{\bsnm{Saska}, \binits{M.}}:
\batitle{{Physical Orienteering Problem for Unmanned Aerial Vehicle Data
  Collection Planning in Environments With Obstacles}}.
\bjtitle{IEEE Robotics and Automation Letters}
\bvolume{4}(\bissue{3}),
\bfpage{3005}--\blpage{3012}
(\byear{2019}).
\doiurl{10.1109/LRA.2019.2923949}
\end{barticle}
\endbibitem

\bibitem{Penicka2022RAL-I}
\begin{barticle}
\bauthor{\bsnm{Penicka}, \binits{R.}},
\bauthor{\bsnm{Scaramuzza}, \binits{D.}}:
\batitle{{Minimum-Time Quadrotor Waypoint Flight in Cluttered Environments}}.
\bjtitle{IEEE Robotics and Automation Letters}
\bvolume{7}(\bissue{2}),
\bfpage{5719}--\blpage{5726}
(\byear{2022}).
\doiurl{10.1109/LRA.2022.3154013}
\end{barticle}
\endbibitem

\bibitem{Penicka2022RAL-II}
\begin{barticle}
\bauthor{\bsnm{Penicka}, \binits{R.}},
\bauthor{\bsnm{Song}, \binits{Y.}},
\bauthor{\bsnm{Kaufmann}, \binits{E.}},
\bauthor{\bsnm{Scaramuzza}, \binits{D.}}:
\batitle{{Learning Minimum-Time Flight in Cluttered Environments}}.
\bjtitle{IEEE Robotics and Automation Letters}
\bvolume{7}(\bissue{3}),
\bfpage{7209}--\blpage{7216}
(\byear{2022}).
\doiurl{10.1109/LRA.2022.3181755}
\end{barticle}
\endbibitem

\bibitem{nguyen2021ECC}
\begin{bchapter}
\bauthor{\bsnm{Nguyen}, \binits{H.}},
\bauthor{\bsnm{Kamel}, \binits{M.}},
\bauthor{\bsnm{Alexis}, \binits{K.}},
\bauthor{\bsnm{Siegwart}, \binits{R.}}:
\bctitle{{Model Predictive Control for Micro Aerial Vehicles: A Survey}}.
In: \bbtitle{European Control Conference},
pp. \bfpage{1556}--\blpage{1563}
(\byear{2021}).
\doiurl{10.23919/ECC54610.2021.9654841}
\end{bchapter}
\endbibitem

\bibitem{robinson2018RAL}
\begin{barticle}
\bauthor{\bsnm{Robinson}, \binits{D.R.}},
\bauthor{\bsnm{Mar}, \binits{R.T.}},
\bauthor{\bsnm{Estabridis}, \binits{K.}},
\bauthor{\bsnm{Hewer}, \binits{G.}}:
\batitle{{An Efficient Algorithm for Optimal Trajectory Generation for
  Heterogeneous Multi-Agent Systems in Non-Convex Environments}}.
\bjtitle{IEEE Robotics and Automation Letters}
\bvolume{3}(\bissue{2}),
\bfpage{1215}--\blpage{1222}
(\byear{2018}).
\doiurl{10.1109/LRA.2018.2794582}
\end{barticle}
\endbibitem

\bibitem{luis2020RAL}
\begin{barticle}
\bauthor{\bsnm{Luis}, \binits{C.E.}},
\bauthor{\bsnm{Vukosavljev}, \binits{M.}},
\bauthor{\bsnm{Schoellig}, \binits{A.P.}}:
\batitle{{Online trajectory generation with distributed model predictive
  control for multi-robot motion planning}}.
\bjtitle{IEEE Robotics and Automation Letters}
\bvolume{5}(\bissue{2}),
\bfpage{604}--\blpage{611}
(\byear{2020}).
\doiurl{10.1109/LRA.2020.2964159}
\end{barticle}
\endbibitem

\bibitem{alcantara2021RAS}
\begin{barticle}
\bauthor{\bsnm{Alc{\'a}ntara}, \binits{A.}},
\bauthor{\bsnm{Capit{\'a}n}, \binits{J.}},
\bauthor{\bsnm{Cunha}, \binits{R.}},
\bauthor{\bsnm{Ollero}, \binits{A.}}:
\batitle{{Optimal trajectory planning for cinematography with multiple unmanned
  aerial vehicles}}.
\bjtitle{Robotics and Autonomous Systems}
\bvolume{140},
\bfpage{103778}
(\byear{2021}).
\doiurl{10.1016/j.robot.2021.103778}
\end{barticle}
\endbibitem

\bibitem{Zhu2019RAL}
\begin{barticle}
\bauthor{\bsnm{Zhu}, \binits{H.}},
\bauthor{\bsnm{Alonso-Mora}, \binits{J.}}:
\batitle{{Chance-Constrained Collision Avoidance for MAVs in Dynamic
  Environments}}.
\bjtitle{IEEE Robotics and Automation Letters}
\bvolume{4}(\bissue{2}),
\bfpage{776}--\blpage{783}
(\byear{2019}).
\doiurl{10.1109/LRA.2019.2893494}
\end{barticle}
\endbibitem

\bibitem{CohenSpringer2023}
\begin{bbook}
\bauthor{\bsnm{Cohen}, \binits{M.}},
\bauthor{\bsnm{Belta}, \binits{C.}}:
\bbtitle{{Adaptive and Learning-Based Control of Safety-Critical Systems}}.
\bpublisher{Springer}, \blocation{???}
(\byear{2023})
\end{bbook}
\endbibitem

\bibitem{Chen2012TRO}
\begin{barticle}
\bauthor{\bsnm{Chen}, \binits{Y.}},
\bauthor{\bsnm{Ding}, \binits{X.C.}},
\bauthor{\bsnm{Stefanescu}, \binits{A.}},
\bauthor{\bsnm{Belta}, \binits{C.}}:
\batitle{{Formal Approach to the Deployment of Distributed Robotic Teams}}.
\bjtitle{IEEE Transactions on Robotics}
\bvolume{28}(\bissue{1}),
\bfpage{158}--\blpage{171}
(\byear{2012}).
\doiurl{10.1109/TRO.2011.2163434}
\end{barticle}
\endbibitem

\bibitem{Lindemann2020TCNS}
\begin{barticle}
\bauthor{\bsnm{Lindemann}, \binits{L.}},
\bauthor{\bsnm{Dimarogonas}, \binits{D.V.}}:
\batitle{{Barrier Function Based Collaborative Control of Multiple Robots Under
  Signal Temporal Logic Tasks}}.
\bjtitle{IEEE Transactions on Control of Network Systems}
\bvolume{7}(\bissue{4}),
\bfpage{1916}--\blpage{1928}
(\byear{2020}).
\doiurl{10.1109/TCNS.2020.3014602}
\end{barticle}
\endbibitem

\bibitem{Gilpin2021LCSS}
\begin{barticle}
\bauthor{\bsnm{Gilpin}, \binits{Y.}},
\bauthor{\bsnm{Kurtz}, \binits{V.}},
\bauthor{\bsnm{Lin}, \binits{H.}}:
\batitle{{A Smooth Robustness Measure of Signal Temporal Logic for Symbolic
  Control}}.
\bjtitle{IEEE Control Systems Letters}
\bvolume{5}(\bissue{1}),
\bfpage{241}--\blpage{246}
(\byear{2021}).
\doiurl{10.1109/LCSYS.2020.3001875}
\end{barticle}
\endbibitem

\bibitem{Leahy2021TRO}
\begin{barticle}
\bauthor{\bsnm{Leahy}, \binits{K.}},
\bauthor{\bsnm{Serlin}, \binits{Z.}},
\bauthor{\bsnm{Vasile}, \binits{C.-I.}},
\bauthor{\bsnm{Schoer}, \binits{A.}},
\bauthor{\bsnm{Jones}, \binits{A.M.}},
\bauthor{\bsnm{Tron}, \binits{R.}},
\bauthor{\bsnm{Belta}, \binits{C.}}:
\batitle{{Scalable and Robust Algorithms for Task-Based Coordination From
  High-Level Specifications (ScRATCHeS)}}.
\bjtitle{IEEE Transactions on Robotics}
\bvolume{38}(\bissue{4}),
\bfpage{2516}--\blpage{2535}
(\byear{2022}).
\doiurl{10.1109/TRO.2021.3130794}
\end{barticle}
\endbibitem

\bibitem{Muniraj2018CDC}
\begin{bchapter}
\bauthor{\bsnm{Muniraj}, \binits{D.}},
\bauthor{\bsnm{Vamvoudakis}, \binits{K.G.}},
\bauthor{\bsnm{Farhood}, \binits{M.}}:
\bctitle{{Enforcing Signal Temporal Logic Specifications in Multi-Agent
  Adversarial Environments: A Deep Q-Learning Approach}}.
In: \bbtitle{IEEE Conference on Decision and Control},
pp. \bfpage{4141}--\blpage{4146}
(\byear{2018}).
\doiurl{10.1109/CDC.2018.8618746}
\end{bchapter}
\endbibitem

\bibitem{RamanCDC2014}
\begin{bchapter}
\bauthor{\bsnm{Raman}, \binits{V.}},
\bauthor{\bsnm{Donzé}, \binits{A.}},
\bauthor{\bsnm{Maasoumy}, \binits{M.}},
\bauthor{\bsnm{Murray}, \binits{R.M.}},
\bauthor{\bsnm{Sangiovanni-Vincentelli}, \binits{A.}},
\bauthor{\bsnm{Seshia}, \binits{S.A.}}:
\bctitle{{Model predictive control with signal temporal logic specifications}}.
In: \bbtitle{53rd IEEE Conference on Decision and Control},
pp. \bfpage{81}--\blpage{87}
(\byear{2014}).
\doiurl{10.1109/CDC.2014.7039363}
\end{bchapter}
\endbibitem

\bibitem{Kantaros2020IJRR}
\begin{barticle}
\bauthor{\bsnm{Kantaros}, \binits{Y.}},
\bauthor{\bsnm{Zavlanos}, \binits{M.M.}}:
\batitle{{STyLuS*: A Temporal Logic Optimal Control Synthesis Algorithm for
  Large-Scale Multi-Robot Systems}}.
\bjtitle{The International Journal of Robotics Research}
\bvolume{39}(\bissue{7}),
\bfpage{812}--\blpage{836}
(\byear{2020}).
\doiurl{10.1177/027836492091392}
\end{barticle}
\endbibitem

\bibitem{Silano2021RAL}
\begin{barticle}
\bauthor{\bsnm{Silano}, \binits{G.}},
\bauthor{\bsnm{Baca}, \binits{T.}},
\bauthor{\bsnm{Penicka}, \binits{R.}},
\bauthor{\bsnm{Liuzza}, \binits{D.}},
\bauthor{\bsnm{Saska}, \binits{M.}}:
\batitle{{Power Line Inspection Tasks With Multi-Aerial Robot Systems Via
  Signal Temporal Logic Specifications}}.
\bjtitle{IEEE Robotics and Automation Letters}
\bvolume{6}(\bissue{2}),
\bfpage{4169}--\blpage{4176}
(\byear{2021}).
\doiurl{10.1109/LRA.2021.3068114}
\end{barticle}
\endbibitem

\bibitem{Bertsekas2012Book}
\begin{bbook}
\bauthor{\bsnm{Bertsekas}, \binits{D.}}:
\bbtitle{{Dynamic Programming and Optimal Control}}.
\bpublisher{Athena Scientific}, \blocation{???}
(\byear{2012}).
\bcomment{{ISBN: 978-3-540-30301-5}}.
\burl{http://www.athenasc.com/dpbook.html}
\end{bbook}
\endbibitem

\bibitem{CaiMacroeconomicDynamics2017}
\begin{barticle}
\bauthor{\bsnm{Cai}, \binits{Y.}},
\bauthor{\bsnm{Judd}, \binits{K.L.}},
\bauthor{\bsnm{Lontzek}, \binits{T.S.}},
\bauthor{\bsnm{Michelangeli}, \binits{V.}},
\bauthor{\bsnm{Su}, \binits{C.-L.}}:
\batitle{{A Nonlinear Programming Method For Dynamic Programming}}.
\bjtitle{Macroeconomic Dynamics}
\bvolume{21}(\bissue{2}),
\bfpage{336}--\blpage{361}
(\byear{2017}).
\doiurl{10.1017/S1365100515000528}
\end{barticle}
\endbibitem

\bibitem{MuellerTRO2015}
\begin{barticle}
\bauthor{\bsnm{Mueller}, \binits{M.W.}},
\bauthor{\bsnm{Hehn}, \binits{M.}},
\bauthor{\bsnm{D'Andrea}, \binits{R.}}:
\batitle{{A Computationally Efficient Motion Primitive for Quadrocopter
  Trajectory Generation}}.
\bjtitle{IEEE Transactions on Robotics}
\bvolume{31}(\bissue{6}),
\bfpage{1294}--\blpage{1310}
(\byear{2015}).
\doiurl{10.1109/TRO.2015.2479878}
\end{barticle}
\endbibitem

\bibitem{Fainekos2009TCS}
\begin{barticle}
\bauthor{\bsnm{{Fainekos}}, \binits{G.E.}},
\bauthor{\bsnm{{Pappas}}, \binits{G.J.}}:
\batitle{{Robustness of temporal logic specifications for continuous-time
  signals}}.
\bjtitle{Theoretical Computer Science}
\bvolume{410}(\bissue{42}),
\bfpage{4262}--\blpage{4291}
(\byear{2009}).
\doiurl{10.1016/j.tcs.2009.06.021}
\end{barticle}
\endbibitem

\bibitem{Souza2007JSTTT}
\begin{barticle}
\bauthor{\bsnm{D'Souza}, \binits{D.}},
\bauthor{\bsnm{Prabhakar}, \binits{P.}}:
\batitle{{On the expressiveness of MTL in the pointwise and continuous
  semantics}}.
\bjtitle{International Journal on Software Tools for Technology Transfer}
\bvolume{9},
\bfpage{1}--\blpage{4}
(\byear{2007}).
\doiurl{10.1007/s10009-005-0214-9}
\end{barticle}
\endbibitem

\bibitem{Pant2017CCTA}
\begin{bchapter}
\bauthor{\bsnm{Pant}, \binits{Y.V.}},
\bauthor{\bsnm{Abbas}, \binits{H.}},
\bauthor{\bsnm{Mangharam}, \binits{R.}}:
\bctitle{{Smooth operator: Control using the smooth robustness of temporal
  logic}}.
In: \bbtitle{2017 IEEE Conference on Control Technology and Applications},
pp. \bfpage{1235}--\blpage{1240}
(\byear{2017}).
\doiurl{10.1109/CCTA.2017.8062628}
\end{bchapter}
\endbibitem

\bibitem{Miller1960}
\begin{barticle}
\bauthor{\bsnm{Miller}, \binits{C.}},
\bauthor{\bsnm{Tucker}, \binits{A.}},
\bauthor{\bsnm{Zemlin}, \binits{R.A.}}:
\batitle{{Integer programming formulation of traveling salesman problems}}.
\bjtitle{Journal of the Association for Computing Machinery}
\bvolume{7},
\bfpage{326}--\blpage{329}
(\byear{1960}).
\doiurl{10.1145/321043.321046}
\end{barticle}
\endbibitem

\bibitem{LaporteNRL2007}
\begin{barticle}
\bauthor{\bsnm{Laporte}, \binits{G.}}:
\batitle{{What you should know about the vehicle routing problem}}.
\bjtitle{Naval Research Logistics}
\bvolume{54}(\bissue{8}),
\bfpage{811}--\blpage{819}
(\byear{2007}).
\doiurl{10.1002/nav.20261}
\end{barticle}
\endbibitem

\bibitem{Laporte1986JOPRS}
\begin{barticle}
\bauthor{\bsnm{Laporte}, \binits{G.}}:
\batitle{{Generalized Subtour Elimination Constraints and Connectivity
  Constraints}}.
\bjtitle{Journal of the Operational Research Society}
\bvolume{37},
\bfpage{509}--\blpage{514}
(\byear{1986}).
\doiurl{10.1057/jors.1986.86}
\end{barticle}
\endbibitem

\bibitem{Yuan2020OPL}
\begin{barticle}
\bauthor{\bsnm{Yuan}, \binits{Y.}},
\bauthor{\bsnm{Cattaruzza}, \binits{D.}},
\bauthor{\bsnm{Ogier}, \binits{M.}},
\bauthor{\bsnm{Semet}, \binits{F.}}:
\batitle{{A note on the lifted Miller–Tucker–Zemlin subtour elimination
  constraints for routing problems with time windows}}.
\bjtitle{Operations Research Letters}
\bvolume{48}(\bissue{2}),
\bfpage{167}--\blpage{169}
(\byear{2020}).
\doiurl{10.1016/j.orl.2020.01.008}
\end{barticle}
\endbibitem

\bibitem{Achuthan1996EJOR}
\begin{barticle}
\bauthor{\bsnm{Achuthan}, \binits{N.R.}},
\bauthor{\bsnm{Caccetta}, \binits{L.}},
\bauthor{\bsnm{Hill}, \binits{S.P.}}:
\batitle{{A new subtour elimination constraint for the vehicle routing
  problem}}.
\bjtitle{European Journal of Operational Research}
\bvolume{91}(\bissue{3}),
\bfpage{573}--\blpage{586}
(\byear{1996}).
\doiurl{10.1016/0377-2217(94)00332-7}
\end{barticle}
\endbibitem

\bibitem{Leishman2006CUP}
\begin{botherref}
\oauthor{\bsnm{Leishman}, \binits{G.J.}}:
{Principles of Helicopter Aerodynamics},
Cambridge University Press
(2006).
{ISBN: 978-1107-0133-53}.
\url{https://tinyurl.com/57tzafdb}
\end{botherref}
\endbibitem

\bibitem{Franco2026JINT}
\begin{barticle}
\bauthor{\bsnm{Carmelo}, \binits{D.}},
\bauthor{\bsnm{Giorgio}, \binits{B.}}:
\batitle{{Coverage Path Planning for UAVs Photogrammetry with Energy and
  Resolution Constraints}}.
\bjtitle{Journal of Intelligent \& Robotic Systems}
\bvolume{83},
\bfpage{445}--\blpage{462}
(\byear{2016}).
\doiurl{10.1007/s10846-016-0348-x}
\end{barticle}
\endbibitem

\bibitem{BauersfeldRAL2022}
\begin{barticle}
\bauthor{\bsnm{Bauersfeld}, \binits{L.}},
\bauthor{\bsnm{Scaramuzza}, \binits{D.}}:
\batitle{{Range, Endurance, and Optimal Speed Estimates for Multicopters}}.
\bjtitle{IEEE Robotics and Automation Letters}
\bvolume{7}(\bissue{2}),
\bfpage{2953}--\blpage{2960}
(\byear{2022}).
\doiurl{10.1109/LRA.2022.3145063}
\end{barticle}
\endbibitem

\bibitem{Baca2020mrs}
\begin{barticle}
\bauthor{\bsnm{{Baca}}, \binits{T.}},
\bauthor{\bsnm{{Petrlik}}, \binits{M.}},
\bauthor{\bsnm{{Vrba}}, \binits{M.}},
\bauthor{\bsnm{{Spurny}}, \binits{V.}},
\bauthor{\bsnm{{Penicka}}, \binits{R.}},
\bauthor{\bsnm{{Hert}}, \binits{D.}},
\bauthor{\bsnm{{Saska}}, \binits{M.}}:
\batitle{{The MRS UAV System: Pushing the Frontiers of Reproducible Research,
  Real-world Deployment, and Education with Autonomous Unmanned Aerial
  Vehicles}}.
\bjtitle{Journal of Intelligent \& Robotic Systems}
\bvolume{102}(\bissue{26}),
\bfpage{1}--\blpage{28}
(\byear{2021}).
\doiurl{10.1007/s10846-021-01383-5}
\end{barticle}
\endbibitem

\bibitem{Silano2019SMC}
\begin{bchapter}
\bauthor{\bsnm{Silano}, \binits{G.}},
\bauthor{\bsnm{Oppido}, \binits{P.}},
\bauthor{\bsnm{Iannelli}, \binits{L.}}:
\bctitle{{Software-in-the-loop simulation for improving flight control system
  design: a quadrotor case study}}.
In: \bbtitle{IEEE International Conference on Systems, Man and Cybernetics},
pp. \bfpage{466}--\blpage{471}
(\byear{2019}).
\doiurl{10.1109/SMC.2019.8914154}
\end{bchapter}
\endbibitem

\bibitem{MRS2022ICUAS_HW}
\begin{bchapter}
\bauthor{\bsnm{{Hert}}, \binits{D.}},
\bauthor{\bsnm{{Baca}}, \binits{T.}},
\bauthor{\bsnm{{Petracek}}, \binits{P.}},
\bauthor{\bsnm{{Kratky}}, \binits{V.}},
\bauthor{\bsnm{{Spurny}}, \binits{V.}},
\bauthor{\bsnm{{Petrilik}}, \binits{M.}},
\bauthor{\bsnm{{Matous}}, \binits{V.}},
\bauthor{\bsnm{{Zaitlik}}, \binits{D.}},
\bauthor{\bsnm{{Stoudek}}, \binits{P.}},
\bauthor{\bsnm{{Walter}}, \binits{V.}},
\bauthor{\bsnm{{Stepan}}, \binits{P.}},
\bauthor{\bsnm{{Horyna}}, \binits{J.}},
\bauthor{\bsnm{{Ptrizl}}, \binits{V.}},
\bauthor{\bsnm{{Silano}}, \binits{G.}},
\bauthor{\bsnm{{Bonilla Licea}}, \binits{D.}},
\bauthor{\bsnm{{Stibinger}}, \binits{P.}},
\bauthor{\bsnm{{Penicka}}, \binits{R.}},
\bauthor{\bsnm{{Nascimento}}, \binits{T.}},
\bauthor{\bsnm{{Saska}}, \binits{M.}}:
\bctitle{{MRS Modular UAV Hardware Platforms for Supporting Research in
  Real-World Outdoor and Indoor Environments}}.
In: \bbtitle{International Conference on Unmanned Aircraft Systems},
pp. \bfpage{1264}--\blpage{1273}
(\byear{2022}).
\doiurl{10.1109/ICUAS54217.2022.9836083}
\end{bchapter}
\endbibitem

\bibitem{MRS2023JINT_HW}
\begin{barticle}
\bauthor{\bsnm{{Hert}}, \binits{D.}},
\bauthor{\bsnm{{Baca}}, \binits{T.}},
\bauthor{\bsnm{{Petracek}}, \binits{P.}},
\bauthor{\bsnm{{Kratky}}, \binits{V.}},
\bauthor{\bsnm{{Penicka}}, \binits{R.}},
\bauthor{\bsnm{{Spurny}}, \binits{V.}},
\bauthor{\bsnm{{Petrilik}}, \binits{M.}},
\bauthor{\bsnm{{Matous}}, \binits{V.}},
\bauthor{\bsnm{{Zaitlik}}, \binits{D.}},
\bauthor{\bsnm{{Stoudek}}, \binits{P.}},
\bauthor{\bsnm{{Walter}}, \binits{V.}},
\bauthor{\bsnm{{Stepan}}, \binits{P.}},
\bauthor{\bsnm{{Horyna}}, \binits{J.}},
\bauthor{\bsnm{{Ptrizl}}, \binits{V.}},
\bauthor{\bsnm{{Sramek}}, \binits{M.}},
\bauthor{\bsnm{{Ahmad}}, \binits{A.}},
\bauthor{\bsnm{{Silano}}, \binits{G.}},
\bauthor{\bsnm{{Bonilla Licea}}, \binits{D.}},
\bauthor{\bsnm{{Stibinger}}, \binits{P.}},
\bauthor{\bsnm{{Nascimento}}, \binits{T.}},
\bauthor{\bsnm{{Saska}}, \binits{M.}}:
\batitle{{MRS Drone: A Modular Platform for Real-World Deployment of Aerial
  Multi-Robot Systems}}.
\bjtitle{Journal of Intelligent \& Robotic Systems}
\bvolume{108}(\bissue{64}),
\bfpage{1}--\blpage{34}
(\byear{2023}).
\doiurl{10.1007/s10846-023-01879-2}
\end{barticle}
\endbibitem

\bibitem{WebbICRA2013}
\begin{bchapter}
\bauthor{\bsnm{Webb}, \binits{D.J.}},
\bauthor{\bparticle{van~den} \bsnm{Berg}, \binits{J.}}:
\bctitle{{Kinodynamic RRT*: Asymptotically optimal motion planning for robots
  with linear dynamics}}.
In: \bbtitle{2013 IEEE International Conference on Robotics and Automation},
pp. \bfpage{5054}--\blpage{5061}
(\byear{2013}).
\doiurl{10.1109/ICRA.2013.6631299}
\end{bchapter}
\endbibitem

\bibitem{Dorling2017IEEESMCS}
\begin{barticle}
\bauthor{\bsnm{Dorling}, \binits{K.}},
\bauthor{\bsnm{Heinrichs}, \binits{J.}},
\bauthor{\bsnm{Messier}, \binits{G.G.}},
\bauthor{\bsnm{Magierowski}, \binits{S.}}:
\batitle{{Vehicle Routing Problems for Drone Delivery}}.
\bjtitle{IEEE Transactions on Systems, Man, and Cybernetics: Systems}
\bvolume{47}(\bissue{1}),
\bfpage{70}--\blpage{85}
(\byear{2017}).
\doiurl{10.1109/TSMC.2016.2582745}
\end{barticle}
\endbibitem

\end{thebibliography}

\end{document}